\documentclass{article}

\usepackage[english]{babel}

\usepackage[a4paper,top=2cm,bottom=2cm,left=2cm,right=2cm,marginparwidth=1.75cm]{geometry}
\everymath{\displaystyle}

\usepackage{amsmath}
\usepackage{amssymb}
\usepackage{mathtools}
\usepackage{bbm}
\usepackage{subfiles} % Best loaded last in the preamble
\usepackage{graphicx}
\usepackage[colorlinks=true, allcolors=blue]{hyperref}
\usepackage{xcolor}
\usepackage{booktabs}
\usepackage{multicol}
\usepackage{multirow}
\usepackage{subcaption}
\usepackage{multimedia}
\usepackage{microtype}
\usepackage{tikz}
\usetikzlibrary{positioning, arrows.meta, fit, shapes.geometric, calc}
\usepackage{caption} % For advanced caption customization
\usepackage{float} % For the float environment

\definecolor{backgroundcolor}{RGB}{245, 246, 246}
\usepackage{tcolorbox}
\tcbuselibrary{listings, skins}
\newtcblisting{pythoncode}[2][]{
    listing only,
    listing options={
        language=Python,
        basicstyle=\ttfamily\small,
        keywordstyle=\color{blue!70},
        commentstyle=\color{green!83!black},
        stringstyle=\color{orange!70},
        numbers=left,
        numberstyle=\tiny\color{black},
        stepnumber=1,
        numbersep=5pt,
        backgroundcolor=\color{backgroundcolor},
        showspaces=false,
        showstringspaces=false,
        showtabs=false,
        frame=none, % Remove the default frame
        tabsize=4,
        captionpos=b,
        breaklines=true,
        breakatwhitespace=true,
        title=\lstname
    },
    colback=backgroundcolor, % Background colour of the box
    colframe=black!80, % Frame color
    sharp corners, % Sharp corners
    boxrule=0.5pt, % Frame thickness
    left=0mm, % Left margin
    right=-1mm, % Right margin
    top=-2mm, % Top margin
    bottom=-3mm, % Bottom margin
    enhanced, % Enable advanced features
    attach boxed title to top left={xshift=5mm, yshift=-1mm}, % Title positioning
    boxed title style={colback=backgroundcolor, colframe=black!80, sharp corners, boxrule=0.5pt}, % Title box style
    drop shadow={black!30}, % Add a shadow
    title=#2, 
    #1
}

\newcommand{\N}{\mathbb{N}}
\newcommand{\Z}{\mathbb{Z}}

\newcommand{\R}{\mathbb{R}}
\newcommand{\C}{\mathbb{C}}

\title{HyperNOs: Automated and Parallel Library for Neural Operators Research}
\date{}
\author{Massimiliano Ghiotto\footnote{Department of Mathematics, University of Pavia, PV 27100, Italy.\\ Email: \href{mailto:massimiliano.ghiotto01@universitadipavia.it}{massimiliano.ghiotto01@universitadipavia.it}, \href{mailto:massimiliano.ghiotto00@gmail.com}{massimiliano.ghiotto00@gmail.com}}}
\begin{document}

% Define a new float environment for code listings
\newfloat{code}{htbp}{loc} % Creates a new float environment called "code"
\floatname{code}{Code} % Sets the name of the float to "Listing"

\maketitle

%%%%%%%%%%%%%%%%%%%%%%%%%%%%%%%%%%%%%%%%%
\begin{abstract}
    This paper introduces HyperNOs, a PyTorch library designed to streamline and automate the process of exploring neural operators, with a special focus on hyperparameter optimization for comprehensive and exhaustive exploration. Indeed, HyperNOs takes advantage of state-of-the-art optimization algorithms and parallel computing implemented in the Ray-tune library to efficiently explore the hyperparameter space of neural operators. We also implement many useful functionalities for studying neural operators with a user-friendly interface, such as the possibility to train the model with a fixed number of parameters or to train the model with multiple datasets and different resolutions. We integrate Fourier neural operators and convolutional neural operators in our library, achieving state of the art results on many representative benchmarks, demonstrating the capabilities of HyperNOs to handle real datasets and modern architectures. The library is designed to be easy to use with the provided model and datasets, but also to be easily extended to use new datasets and custom neural operator architectures.
\end{abstract}

%%%%%%%%%%%%%%%%%%%%%%%%%%%%%%%%%%%%%%%%%
\section{Introduction}

In recent years, the field of scientific machine learning has witnessed significant advances, particularly in the development of methods for solving partial differential equations (PDEs) and modelling complex dynamical systems. Traditional numerical methods, such as finite element analysis, isogeometric analysis \cite{iga05hughes} or spectral methods \cite{spectral06canuto}, have long been the cornerstone of computational science and engineering. However, these methods often face challenges in scalability, computational cost, and generalization across different problem settings. In this context, Neural Operators (NOs) have emerged as a powerful paradigm that provides a data-driven approach to learning mappings between infinite-dimensional function spaces, allowing PDEs and other operator-related problems to be solved with unprecedented efficiency and flexibility.

Neural operators, introduced as a generalization of neural networks to infinite-dimensional spaces, are designed to approximate operators that map input functions to output functions. Unlike traditional neural networks, which operate on finite-dimensional vectors, neural operators are capable of handling continuous functions, making them particularly suited for tasks such as solving PDEs, modelling physical systems, and performing surrogate modelling in high-dimensional spaces. The foundational work on neural operators, including architectures such as Deep Operator Networks (DeepONets, \cite{DON21lu}), Graph-based Neural Operators (GNOs, \cite{GNO20li}),  and Fourier Neural Operators (FNOs, \cite{FNO20li}),  has demonstrated their ability to learn solution operators for a wide range of PDEs. Many modifications of the above architectures have been progressively proposed, such as NOMAD \cite{NOMAD22seidman} as a DeepONet modification, Multiple Graph Neural Operator \cite{MGNO20li} as a GNO modification, and many FNO modifications such as \cite{FFNO21tran, UFNO22wen, OPNO24liu, GIFNO24li}. New architectures are also proposed, such as the Spectral Neural Operator (SNO, \cite{SNO23fanaskov}), the Convolutional Neural Operator (CNO, \cite{CNO23raonic}), and transformer-based operator learning \cite{transformer21cao, poseidon24her}.

The key advantage of neural operators lies in their ability to generalize across different discretizations of the domains, making them highly versatile for various applications, such as weather forecasting \cite{weatherFNO22pathak} or seismic waves \cite{seismicFNO23li}. By leveraging techniques from functional analysis, harmonic analysis, and deep learning, neural operators can be trained on data generated from traditional numerical PDE approximation or with physical measures and subsequently used to predict solutions for new inputs. Moreover, it is theoretically demonstrated (see \cite{don22lan, don23mar} for DeepONets, \cite{uniFNO21kov} for FNOs, \cite{CNO23raonic} for CNOs and \cite{nno23lan} for NOs in general), that neural operators can approximate the solution operators for many families of PDEs, enabling rapid and accurate predictions for new instances of the problem. This capability is particularly valuable in scenarios where traditional numerical methods are computationally expensive or infeasible, such as for high-dimensional problems, many query problems (e.g., uncertainty quantification or inverse problems) and real-time simulations. The versatility of neural operators has led to their adoption in a wide range of scientific applications and engineering domains.

Despite their impressive capabilities, training neural operators presents several challenges. One of the primary challenges is the selection of appropriate hyperparameters, which play a critical role in determining the performance and generalization ability of the model. Hyperparameters include architectural choices (e.g., the number of layers, the type of activation functions, and the size of the hidden layers), optimization parameters (e.g., learning rate, batch size, and optimizer type), and regularization parameters (e.g., learning rate scheduler and weight decay). Because the hyperparameter space like grid search is often large and complex, manual hyperparameter tuning is a labour-intensive and error-prone process that often requires domain expertise, and banal search spaces are too computationally expensive. Automated hyperparameter tuning is a fundamental component of modern machine learning workflows, especially for complex models such as neural operators. The process involves systematically searching the hyperparameter space to identify configurations that maximize model performance, as measured by a predefined loss function. Automated hyperparameters tuning provides several key benefits:
\begin{itemize}
    \item \textbf{Improved Model Performance:} The performance of machine learning models, especially neural operators, is highly sensitive to the choice of hyperparameters. Automated tuning identifies configurations that maximize model performance on validation data, leading to better generalization and robustness. This is particularly important for neural operators, where the relationship between hyperparameters and performance is often non-intuitive and problem-dependent.
    \item \textbf{Efficiency and Automation:} Automated tuning algorithms systematically explore the hyperparameter space, reducing the need for manual intervention. Manual hyperparameter tuning is time-consuming and labour-intensive, requiring extensive trial and error and some domain knowledge, this automation saves time and allows researchers to focus on higher-level tasks, such as model design and interpretation of results.
    \item \textbf{Adaptability to Different Domains:} Neural operators are used in a wide variety of domains and applications, each with its own unique challenges and requirements. Automated tuning can adapt to these diverse contexts by optimizing hyperparameters for specific tasks, datasets, and performance metrics.
    \item \textbf{Scalable and Parallel:} As models and datasets grow in size and complexity, the hyperparameter optimization cost grows exponentially. Manual tuning becomes impractical for large problems due to the sheer number of possible configurations. Automated tuning methods efficiently handle large and high-dimensional search spaces using parallel computing and concentrating on promising hyperparameters configuration and automatically stops bad trials.
    \item \textbf{Reproducibility:} Reproducibility is a cornerstone of scientific research, ensuring that results can be independently verified and validated. We provide a systematic approach that stores hyperparameter configurations and trained models to ensure reproducible results.
    \item \textbf{Human and Machine Efficiency:} As mentioned above, human intuition is limited in navigating complex hyperparameter spaces, especially for high-dimensional models such as neural operators. We show numerically that by offloading the tuning process to machines, trained models are not affected by local minima due to suboptimal hyperparameter choices.
\end{itemize}

To address this challenge, we introduce \textbf{HyperNOs}, a PyTorch library designed to streamline and automate the process of exploring neural operators, with a special focus on hyperparameter optimization for comprehensive and exhaustive exploration. HyperNOs takes advantage of state-of-the-art optimization algorithms and parallel computing implemented in the Ray-tune library to efficiently explore the hyperparameter space of neural operators. We also implement many useful functionalities for studying neural operators with a user-friendly interface. Furthermore, we provide a set of code examples with different datasets and different neural operator architectures, in particular with Fourier neural operators and convolutional neural operators, to demonstrate the capabilities of HyperNOs to handle real datasets and modern architectures. For each example, we provide the associated optimal hyperparameter configuration found and provide the final trained model for future testing. The library is designed to be easy to use with the provided model and datasets, but also to be easily extended to use new datasets and custom neural operator architectures.

In Section \ref{section:neural_operators} we provide an overview of neural operators in general, and then we resume the definition of FNOs and CNOs architectures, highlighting their similarities and differences. In Section \ref{section:hypernos_model_problem} we present the design and main features of the HyperNOs library, for the study of neural operators, through an example of an application. In Section \ref{section:advanced_examples} we show how to use the HyperNOs library with more advanced features that are implemented, such as finding hyperparameters with a given and fixed complexity, enabling multiresolution training or multiple datasets training. In Section \ref{section:numerical_experiments} we provide some numerical experiments to demonstrate the capabilities of the HyperNOs library. Finally, in Section \ref{section:conclusions} we provide some conclusions and future developments.

%%%%%%%%%%%%%%%%%%%%%%%%%%%%%%%%%%%%%%%%%
\section{Overview on Neural Operators}\label{section:neural_operators}
Neural operators have emerged as a powerful framework for learning mappings between infinite dimensional function spaces, making them particularly well suited to solving partial differential equations (PDEs); indeed, neural operators can capture complex nonlinear dependencies and provide efficient approximations. One of the key advantages of neural operators is their ability to generalise beyond the training data. Once trained, a neural operator can provide solutions to a family of PDEs without the need for retraining, significantly reducing the computational cost compared to traditional solvers. This property is particularly beneficial in scenarios where real-time predictions or rapid exploration of parameter spaces are required.

\subsection{Problem Formulation}
We consider a generic family of partial differential equations (PDEs) defined in a bounded domain \( D \subset \mathbb{R}^d \) with boundary \( \partial D \). The PDEs are of the form:
\[
    \begin{cases}
        (\mathcal{L}_a u)(x) = f(x), \quad & x \in D,          \\
        (\mathcal{B}u)(x)    = g(x),       & x \in \partial D,
    \end{cases}
\]
where:
\begin{itemize}
    \item \( \mathcal{L}_a \) is a differential operator parameterized by \( a \in \mathcal{A} \),
    \item \( u: D \to \mathbb{R} \) is the solution to the PDE,
    \item \( f: D \to \mathbb{R} \) is a forcing term or source function.
    \item \(\mathcal{B}\) is possibly a differential operator that with \( g: \partial D \to \mathbb{R} \)  define the boundary condition.
\end{itemize}
\noindent
The parameter \( a \) belongs to the Banach space \( \mathcal{A} \), the solution \( u \) lives in the Banach space \( \mathcal{U}\) and the forcing term \( f \) belongs to the dual space \( \mathcal{U}^* \), which consists of functionals acting on \( \mathcal{U} \). The operator \( \mathcal{L}_a \) maps the parameter \( a \) with a possibly non-linear differential operator \( \mathcal{L}_a: \mathcal{U} \to \mathcal{U}^* \). The solution operator \( \mathcal{G}^\dagger := \mathcal{L}_a^{-1}f \) is defined as the mapping from the parameter \( a \) to the solution \( u \), i.e., \( \mathcal{G}^\dagger: \mathcal{A} \to \mathcal{U} \). This operator is central to our problem, as it encapsulates the relationship between the parameter \( a \) and the solution \( u \). Our goal is to approximate \( \mathcal{G}^\dagger \) using data-driven methods.
\\
In the previous example, we considered the case of a PDE with a single parameter that is \( a \) and fixed forcing term and boundary conditions, but the framework can be trivially modified to consider the forcing term or the boundary condition as the parameter and take \(a\) as a constant. Furthermore, the framework can be easily extended to consider multiple parameters at the same time, i.e., varying both source terms, boundary conditions and the geometry of the problem \cite{moller24iganets}.

\subsection{Data and discretization}\label{subsection:data_discretization}
To make the problem computationally tractable, we assume that the domain \( D \) is discretized into \( n_{points} \in \N \) points. Let \( \{a^{(i)}, u^{(i)}\}_{i=1}^N \) be a dataset of \( N \in \N \) pairs of coefficient functions and solution functions, where \( a^{(i)} \) are i.i.d. samples from a probability measure \( \mu \) supported on \( \mathcal{A} \), and \( u^{(i)} = \mathcal{G}^\dagger(a^{(i)}) \) are the corresponding solutions.
\\
The dataset \( \{a^{(i)}, u^{(i)}\}_{i=1}^N \) is used to train a neural operator model, $\mathcal{G}_{\theta}: \mathcal{A} \to \mathcal{U}$, that approximates the solution operator \( \mathcal{G}^\dagger \). We are interested in controlling the error of the approximation on average with respect to $\mu$, more precisely we want to approximate the solution of the following problem:
\[
    \theta^* = \underset{\theta\in\Theta}{argmin} \ \mathbb{E}_{a\sim\mu}\ Loss\left(\mathcal{G}^\dagger(a), \mathcal{G}_\theta(a)\right) \approx \underset{\theta\in\Theta}{argmin}\ \frac{1}{N} \sum_{i=1}^N \, Loss\left(u^{(i)}, \mathcal{G}_\theta(a^{(i)})\right),
\]
where \( Loss \) is a suitable loss function, and \( \Theta \) is the set of trainable parameters of the neural operator model. A standard choice for the loss function is the relative $L^p$ error between the true solution and the predicted solution
\[
    Loss(u, \mathcal{G}_\theta(a)) = \frac{\|u - \mathcal{G}_\theta(a)\|_{L^p(D)}}{\|u\|_{L^p(D)}} \approx \frac{\left(\sum_{j=1}^{n_{points}} |u(x_j) - \mathcal{G}_\theta(a)(x_j)|^p\right)^{1/p}}{\left(\sum_{j=1}^{n_{points}} |u(x_j)|^p\right)^{1/p}}, \quad p\ge 1.
\]
The final discretized problem is to approximate the parameters of the model that satisfy the following minimization problem
\[
    \theta^* \approx \underset{\theta\in\Theta}{argmin} \ \frac{1}{N} \sum_{i=1}^N \, \frac{\left(\sum_{j=1}^{n_{points}} |u^{(i)}(x_j) - \mathcal{G}_\theta(a^{(i)})(x_j)|^p\right)^{1/p}}{\left(\sum_{j=1}^{n_{points}} |u^{(i)}(x_j)|^p\right)^{1/p}}.
\]
We underline that we need to discretize the problem in order to compute it numerically, but a key feature of neural operators is that they can learn the solution operator in function spaces, so the model definition is independent of the resolution of the discretization of the domain. We mention that exist other loss functions that can be used, e.g., the $H^p$ relative error. For our exposition, and for the neural operator models that we will consider in the coming sections, the previous discretization is natural, but we note that other types of discretization, with appropriate neural operator architecture, can be developed, for example using truncated basis coefficients \cite{SNO23fanaskov}.

\subsection{Multi-Layer Perceptron}
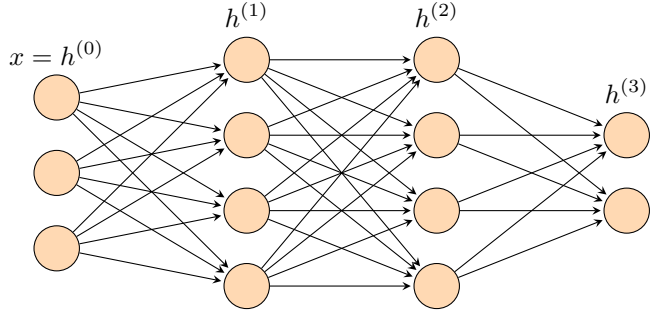
\begin{figure}[!ht]
    \centering
    \begin{tikzpicture}[
            >=stealth,
            shorten >=1pt,
            auto,
            node distance=2cm,
            neuron/.style={draw, circle, fill=orange!30, minimum size=17pt, inner sep=0pt},
            layer/.style={thick, draw=black!50, rectangle, fill=black!10, minimum height=4em},
        ]

        % Draw the input layer nodes
        \foreach \name / \y in {1}
        \node[neuron, label=above:{$x = h^{(0)}$}] (I-\name) at (0,-\y-0.5) {};
        \foreach \name / \y in {2,...,3}
        \node[neuron] (I-\name) at (0,-\y-0.5) {};

        % Draw the hidden layer nodes
        \foreach \name / \y in {1}
        \node[neuron, label=above:{$h^{(1)}$}] (H1-\name) at (2.5,-\y) {};
        \foreach \name / \y in {2,...,4}
        \node[neuron] (H1-\name) at (2.5,-\y) {};

        \foreach \name / \y in {1}
        \node[neuron, label=above:{$h^{(2)}$}] (H2-\name) at (5,-\y) {};
        \foreach \name / \y in {2,...,4}
        \node[neuron] (H2-\name) at (5,-\y) {};

        % Draw the output layer node
        \foreach \name / \y in {2}
        \node[neuron, label=above:{$h^{(3)}$}] (O-\name) at (7.5,-\y) {};
        \foreach \name / \y in {3}
        \node[neuron] (O-\name) at (7.5,-\y) {};

        % Connect every node in the input layer with every node in the
        % hidden layer.
        \foreach \source in {1,...,3}
        \foreach \dest in {1,...,4}
        \draw[-stealth] (I-\source) -- (H1-\dest);

        \foreach \source in {1,...,4}
        \foreach \dest in {1,...,4}
        \draw[-stealth] (H1-\source) -- (H2-\dest);

        % Connect every node in the hidden layer with the output layer
        \foreach \source in {1,...,4}
        \foreach \dest in {2,3}
        \draw[-stealth] (H2-\source) -- (O-\dest);
    \end{tikzpicture}%
    \caption{Representation of a MLP with $L=2$ hidden layers.}
    \label{fig:mlp}
\end{figure}

We have to briefly define \textbf{Multi-Layer Perceptron (MLP)} as a class of feedforward neural networks consisting of multiple layers of neurons, where each neuron applies a nonlinear activation function to its inputs. Mathematically, an MLP with \( L \) layers can be defined as follows. Let \( x \in \R^{n_0} \) be the input vector, and let \( h^{(l)} \in \R^{n_l} \) denote the output of the \( l \)-th layer for \( l = 1, 2, \dots, L+1 \). The MLP is defined by the following recursive relation:
\[
    \begin{cases}
        h^{(l)} = \sigma\left(W^{(l)} h^{(l-1)} + b^{(l)}\right), \quad l = 1, 2, \dots, L, \\
        h^{(L+1)} = W^{(L+1)} h^{(L)} + b^{(L+1)},
    \end{cases}
\]
where \( h^{(0)} := x \) is the input to the first layer, \( W^{(l)} \in \R^{n_l \times n_{l-1}}\), for \( l = 1, \dots, L+1 \), is the weight matrix of the \( l \)-th layer, \( b^{(l)} \in \R^{n_l} \), for \( l = 1 , \dots, L+1 \), is the bias vector of the \( l \)-th layer, \( \sigma: \R \to \R \) is a nonlinear activation function applied element-wise and  $h^{(L+1)}\in \R^{n_{L+1}}$, is the final output vector. You can see a visual representation of an MLP with \( L = 2 \) hidden layers in Figure \ref{fig:mlp}.

\subsection{Neural Operators}
In this section, we provide a general overview of neural operators. In particular, we try to give a general definition of neural operators, following \cite{neuraloperator21kov}, highlighting the main properties that a neural operator needs to satisfy. We then specify the NO definition to the FNO and CNO models.
\\
We define a neural operator as a composition of other operators as follows:
\[ \mathcal{G}_{\theta} :\mathcal{A}(D, \R^{d_a}) \to \mathcal{U}(D, \R^{d_{u}}), \quad \text{with } D \subset \R^d, \quad	\mathcal{G}_{\theta} := \mathcal{Q} \circ \mathcal{L}_L \circ \cdots \circ \mathcal{L}_1 \circ \mathcal{P} . \]
\begin{enumerate}
    \item \textbf{Lifting operator:} The lifting operator is the first step in the neural operator architecture. It maps the input function from its original function space $(d_a)$ to a higher-dimensional space $(d_v)$. This is useful because the input function is often low-dimensional (e.g., scalar-valued), while the neural operator requires a higher-dimensional representation to effectively process and extract features. The lifting operator is defined as follows:
          \[\mathcal{P}:\, \mathcal{A}(D, \R^{d_a}) \to \mathcal{U}(D, \R^{d_{v}}), \quad  v_0(x) = \mathcal{P}(a)(x) = \mathcal{P}(a(x)) \in \R^{d_v}.\]
          The lifting operator is typically implemented as an MLP (defined above) or a convolution block (defined later in Section \ref{subsection:cno}) that acts locally on the input function. Generally, it is a shallow network or even a linear layer, respectively, an MLP (or convolution block) with one and zero hidden layers.
    \item \textbf{Integral operator:} The integral operator is the core component of the neural operator architecture. It captures the global interactions and dependencies in the input function by applying a sequence of linear and non-linear transformations. The integral operator is composed of \( L \) layers, each of which is as follows
          \[ \mathcal{L}_t : \, \mathcal{U}(D, \R^{d_{v}}) \to  \mathcal{U}(D, \R^{d_{v}}), \quad t = 1, \dots, L. \]
          The resulting operator is a composition of \( L \) layers, and this composition has to be non-linear and non-local in order to approximate the solution operator coming from the PDEs.
    \item \textbf{Projection operator:} The projection operator is the final step in the neural operator architecture. It maps the high-dimensional internal representation of the network ($d_v$) back to the output function space ($d_u$).
          \[\mathcal{Q}:\, \mathcal{U}(D, \R^{d_{v}}) \to  \mathcal{U}(D, \R^{d_{u}}), \quad  \mathcal{Q}(v_L)(x) = \mathcal{Q}(v_L(x)) = u(x) \in \R^{d_u}. \]
          The projection operator acts pointwise and can be non-linear, the implementation is the same for the lifting operator.
\end{enumerate}

\subsection{Fourier Neural Operators}\label{subsection:fno}
Fourier neural operators are a class of neural operators that use the Fourier transform to efficiently parameterize the integral operator defined in the general neural operator framework. As the Fourier transform is a global operator, each layer of the neural operator is non-local and is defined as non-linear by composition with a non-linear activation function. By operating in the frequency domain, FNOs can capture global dependencies in the input functions, making them highly effective for problems with periodic boundary conditions or smooth solutions.

More precisely the integral operator for FNOs is defined as follows:
\begin{equation}
    v_{t+1}(x) = \mathcal{L}_t(v_t)(x) := \sigma\Big( W_t v_t(x)+ b_t + (\mathcal{K}_t(a, \theta_t) v_t)(x) \Big), \quad W_{t} \in \R^{d_{v} \times d_v},\ b_{t} \in \R^{d_v}, \quad t = 0, \dots, L-1,
    \label{eq:integral_operator_fno}
\end{equation}
$\theta_t \in \Theta_t \subset \Theta$ is a subset of all the trainable parameters, $\sigma$ is a non-linear activation function. Moreover, $\mathcal{K}_t(a, \theta_t)$ is called the \textbf{integral kernel operator}. The first two addends in \eqref{eq:integral_operator_fno} is a pointwise and linear transformation, whereas the integral kernel operator is a linear but non-local operator. Moreover, for the definition of the Fourier neural operator, we make the following assumptions
\[ (\mathcal{K}_t(a, \theta_t)v_t)(x) = (\mathcal{K}_t(\theta_t)v_t)(x) = \int_{D} \kappa_{t,\theta_t}(x,y) v_t(y) \ dy = \int_{D} \kappa_{t,\theta_t}(x-y) v_t(y) \ dy= (\kappa_{t, \theta_t} * v_t)(x) ,\]
where we assume $D = \mathbb{T}^d$ that is the $d$-dimensional torus, $\kappa_{t,\theta_t}(z) \in\R^{d_v\times d_v}, z \in D,$ is a learnable function usually called kernel function, and $*$ denotes the convolution of functions. In FNOs, the kernel is parameterized in the Fourier domain, allowing for efficient computation and global feature extraction, indeed using the convolution theorem we have
\[ (\kappa_{t, \theta_t} * v_t)(x) =  \mathcal{F}^{-1}\left( \mathcal{F}( \kappa_{t,\theta_t}) (k) \cdot \mathcal{F}(v_t)(k) \right)(x), \]
where $\mathcal{F}$, $\mathcal{F}^{-1}$ are respectively the Fourier transform and the inverse Fourier transform. Furthermore $\mathcal{F}( \kappa_{t,\theta_t}) (k) \in \C^{d_v \times d_v}$, $\mathcal{F}(v_t)(k)\in \C^{d_v}$ for every $k \in \Z^d$ and the multiplication dot represents the standard matrix-vector multiplication. Parameterizing $\mathcal{F}( \kappa_{t, \theta_t} )(k)$ with learnable parameters $R_{\theta_t}(k) \in \C^{d_v \times d_v}, \ \forall \ k \in \Z^d $, we obtain
\begin{equation}
    (\mathcal{K}_t(\theta_t)v_t)(x)= \mathcal{F}^{-1}\left( R_{\theta_t}(k) \cdot \mathcal{F}(v_t)(k) \right)(x).
    \label{eq:discretized_integral_operator_fno}
\end{equation}
Since the function $k_{t, \theta_t}$ is a real-valued function that is parameterized by $R_{\theta_t}(k),\ k \in \Z^d, $ we have to impose that the parameters satisfy the Hermitian property, i.e., $R_{\theta_t}(k) = \overline{R_{\theta_t}}(-k),\ \forall \ k \in \Z^d$.

\begin{figure}[!ht]
    \centering
    \begin{tikzpicture}
        \tikzset{
            box/.style={draw, rounded corners, align=center, minimum height=0.8cm, minimum width=1cm, fill=orange!30},
            bigbox/.style={draw, rounded corners, align=center, minimum height=2cm, minimum width=1cm, fill=yellow!30},
            node_sum/.style={draw, circle, fill=white, inner sep=0pt, minimum size=4mm},
            every node/.style={font=\small}
        }
        % Nodes for the main structure
        \node[box, label=above:{\textit{Input}}] (input) {$a(x)$};
        \node[box, right=0.5cm of input, label=above:{\textit{Lifting}}] (lifting) { $\mathcal{P}$ };
        \node[bigbox, right=0.5cm of lifting] (fourier1) {$\mathcal{L}_{1}$};
        \node[bigbox, right=0.55cm of fourier1] (fourier2) {$\mathcal{L}_{t}$};
        \node[bigbox, right=0.55cm of fourier2] (fourier3) {$\mathcal{L}_{L}$};
        \node[box, right=0.5cm of fourier3, label=above:{\textit{Projection}}] (projection) {$\mathcal{Q}$};
        \node[box, right=0.5cm of projection, label=above:{\textit{Output}}] (output) {$u(x)$};

        % Draw arrows between nodes
        \draw[-stealth, line width = .7pt] ($(input.east)+(0.05, 0)$) -- ($(lifting.west)-(0.05,0)$);
        \draw[-stealth, line width = .7pt] ($(lifting.east)+(0.05, 0)$) -- ($(fourier1.west)-(0.03,0)$);
        \draw[dotted, line width = 2pt] ($(fourier1.east)+(0.1, 0)$) -- ($(fourier2.west)-(0.1,0)$);
        \draw[dotted, line width = 2pt] ($(fourier2.east)+(0.1, 0)$) -- ($(fourier3.west)-(0.1,0)$);
        \draw[-stealth, line width = .7pt] ($(fourier3.east)+(0.05, 0)$) -- ($(projection.west)-(0.05,0)$);
        \draw[-stealth, line width = .7pt] ($(projection.east)+(0.05, 0)$) -- ($(output.west)-(0.05,0)$);

        % Annotations for the Fourier Layers
        \node[align=center, above=0.2cm of fourier2, font=\footnotesize] { \textit{Fourier Layers} };

        % Internal structure bounding box
        \node[draw, line width = .7pt, rounded corners, inner sep=0.2cm, fit= (fourier1) (fourier2) (fourier3)] (internal) {};

        %%% Second part of the plot
        % Internal structure bounding box
        \node[draw, below=0.2cm of internal, rounded corners, inner sep=0.2cm, fill=yellow!30] (internal) {
            \begin{tikzpicture}[every node/.style={font=\small}]
                % Nodes for the internal structure
                \node[box] (vt) {$v_t(x)$};
                \node[box, right=1cm of vt] (transform) {$\mathcal{F}$};
                \node[box, right=0.5cm of transform, fill=green!25] (nonlinear) { $ R_{\theta_t} $ };
                \node[box, right=0.5cm of nonlinear] (invtransform) { $ \mathcal{F}^{-1} $ };
                \node[box, below=0.8cm of nonlinear, fill=green!25] (linear) { $ W_t, b_t $ };

                % Internal structure bounding box
                \node[draw, line width = .7pt, rounded corners, inner sep=0.15cm, fit= (transform) (nonlinear) (invtransform) ] (diagonalscaling) {};

                \node[node_sum, right=0.5cm of invtransform] (node_sum) {$\mathbf{+}$};
                \node[box, right=0.5cm of node_sum] (activation) {$\sigma$};
                \node[box, right=0.5cm of activation] (vtplusone) {$v_{t+1}(x)$};

                % Draw arrows between nodes
                \draw[line width = .7pt] ($(vt.east)+(0.05, 0)$) -- (diagonalscaling);
                \draw[-stealth, line width = .7pt] ($(transform.east)+(0.05, 0)$) -- ($(nonlinear.west)-(0.05, 0)$);
                \draw[-stealth, line width = .7pt] ($(nonlinear.east)+(0.05, 0)$) -- ($(invtransform.west)-(0.05, 0)$);
                \draw[-stealth, line width = .7pt] ($(vt.south)-(0, 0.05)$) |- ($(linear.west)-(0.05, 0)$);
                \draw[-stealth, line width = .7pt] (diagonalscaling) -- ($(node_sum.west)-(0.05, 0)$);
                \draw[-stealth, line width = .7pt] ($(linear.east)+(0.05, 0)$) -| ($(node_sum.south)-(0, 0.05)$);
                \draw[-stealth, line width = .7pt] ($(node_sum.east)+(0.05, 0)$) -- ($(activation.west)-(0.05, 0)$);
                \draw[-stealth, line width = .7pt] ($(activation.east)+(0.05, 0)$) -- ($(vtplusone.west)-(0.05, 0)$);

            \end{tikzpicture}
        };
        % Connection between the two parts
        \draw[] ($(internal.north west)+(0.1, 0.05)$) -- (fourier2.south west);
        \draw[] ($(internal.north east)+(-0.1, 0.05)$) -- (fourier2.south east);
    \end{tikzpicture}
    \caption{Visual representation of a Fourier Neural Operator.}
    \label{fig:fno}
\end{figure}
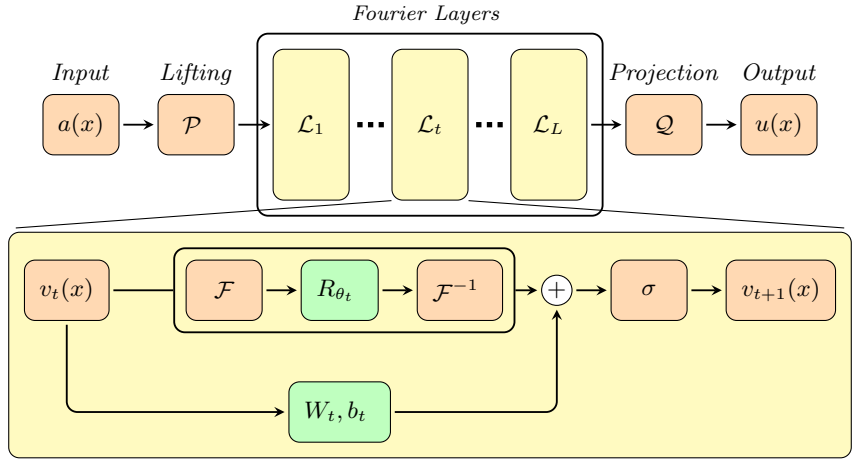

To simplify the understanding of the Fourier we provide Figure \ref{fig:fno} for a visualization of the model architecture, for a video representation see \cite{Max25FNOvideo}.
As a final remark, we mention that in practice some slight modifications of \eqref{eq:integral_operator_fno} are considered. In particular, we introduce the following three notations:
\[
    \begin{aligned}
        v_{t+1}^{Classic} & = \mathcal{L}_t^{Classic}(v_t) := \sigma\Big( W_t v_t+ b_t + \mathcal{K}_t(\theta_t) v_t \Big),  \\
        v_{t+1}^{MLP}     & = \mathcal{L}_t^{MLP}(v_t) := \sigma\Big( W_t v_t+ b_t + MLP(\mathcal{K}_t(\theta_t) v_t )\Big), \\
        v_{t+1}^{Res}     & = \mathcal{L}_t^{Res}(v_t) := v_t+\sigma\Big( W_t v_t+ b_t + \mathcal{K}_t(\theta_t) v_t \Big).
    \end{aligned}
\]
Where the first one is exactly \eqref{eq:integral_operator_fno} with a different notation, the second one adds an MLP acting pointwise after the integral kernel operator and the third one is a residual-like modification. It is numerically shown that employing skip connections to Fourier layers enables the training of deeper FNOs \cite{FFNO21tran}.

\subsection{Convolutional Neural Operators}\label{subsection:cno}

Convolutional neural operators extend the concept of Convolutional Neural Networks (CNNs), in particular U-Nets \cite{unet15ron}, to the setting of neural operators. As studied in \cite{reno24bar} a Representative Equivalent Neural Operator (RENO) have to satisfy the Continuous-Discrete Equivalence (CDE) in order to learn the underlying operator and not only a discrete representation of it. This feature is crucial for neural operators, to ensure stable model performance across different discretization resolutions. The independence of the architecture from the discretization resolution (also called resolution invariance property) is necessary but not sufficient to be RENO. Indeed, CNOs are carefully designed to satisfy the resolution invariant and the CDE condition. CNOs are defined on the space of bandlimited functions, defined as
\[
    \mathcal{B}_{\omega}(D, \R^{d_v}) = \{f \in L^2(D, \R^{d_v}) : supp\mathcal{F}(f) \subseteq [-\omega, \omega]^2 \},
\]
where $\omega>0$,  $\mathcal{F}(f)$ denotes the Fourier transform of the function $f$ and until the end of this section we assume $D$ a rectangular domain. From the definition of the convolutional neural operator will follows that CNO maps from the space of bandlimited functions to the space of bandlimited functions, i.e., $\mathcal{G}_{\theta}:\mathcal{B}_{\omega}(D, \R^{d_a}) \to \mathcal{B}_{\omega}(D, \R^{d_u})$.
\\
With the previous settings, we define the lifting and projection operator as a convolutional block (defined later in the section) and the integral operator for CNOs as follows:
\begin{equation}
    v_{t+1} = \mathcal{L}_t(v_t) := \mathcal{P}_t \circ \Sigma_t \circ \mathcal{K}_t(v_t), \quad  t = 0, \dots, L-1.
    \label{eq:integral_operator_cno}
\end{equation}
with $v_{t}, v_{t+1} \in \mathcal{B}_{\omega}(D, \R^{d_v})$. As we can see from \eqref{eq:integral_operator_cno}, the lifted function $(v_0)$ is processed through the composition of a series of mappings between functions (layers), with each layer consisting of three elementary operators, i.e., $\mathcal{P}_t$ is either the \textit{upsampling} or \textit{downsampling} operator, $\mathcal{K}_t$ is the convolution operator and $\Sigma_t$ is the activation operator. In the remaining of this section where are going to describe these three elementary operators. Furthermore, to construct the typical U-Net architecture, in addition to the previously mentioned operators, we will define the residual operator and the invariant operator.

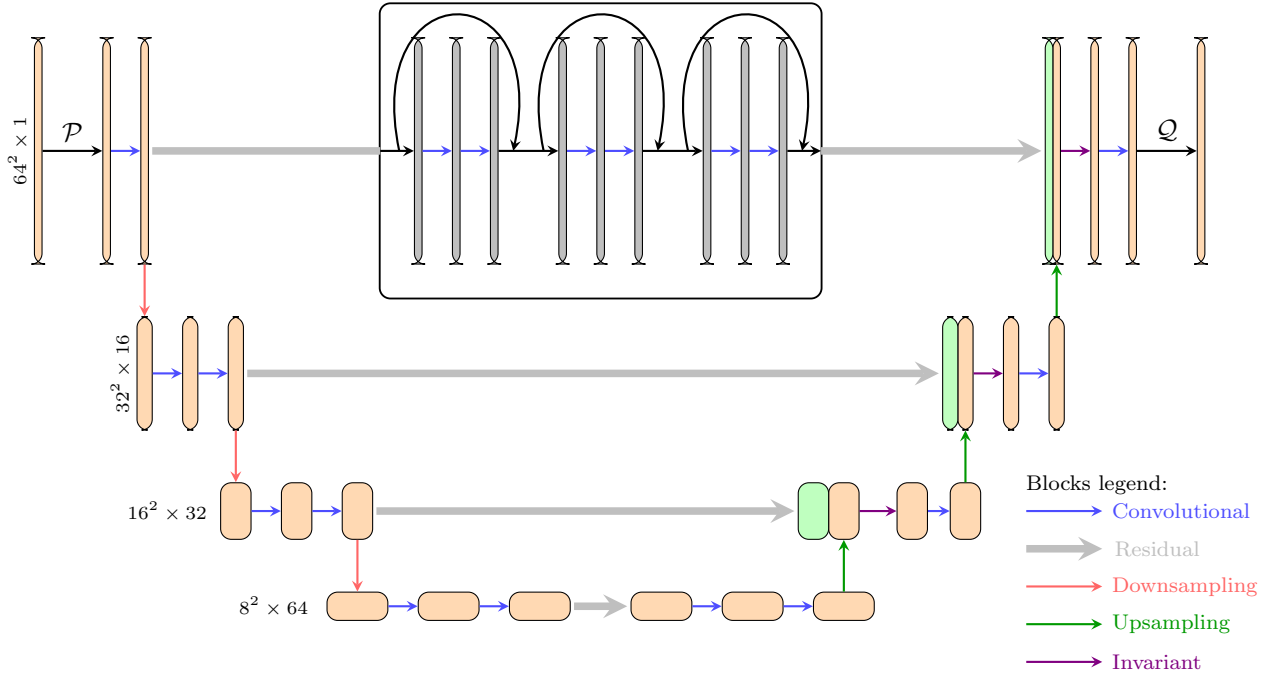
\begin{figure}[!ht]
    \centering
    \begin{tikzpicture}[
            % Styles
            box/.style n args={2}{
                    draw,
                    rounded corners,
                    align=center,
                    minimum height=#1,
                    minimum width=#2,
                    text width=#2,
                    inner sep=0pt,
                    fill=orange!30
                },
            greenbox/.style n args={2}{
                    draw,
                    rounded corners,
                    align=center,
                    minimum height=#1,
                    minimum width=#2,
                    text width=#2,
                    inner sep=0pt,
                    fill=green!25
                },
            graybox/.style n args={2}{
                    draw,
                    rounded corners,
                    align=center,
                    minimum height=#1,
                    minimum width=#2,
                    text width=#2,
                    inner sep=0pt,
                    fill=gray!50
                },
            conv/.style={-stealth, blue!70, thick},
            inv/.style={-stealth, violet, thick},
            connection/.style={-stealth, black, thick},
            pool/.style={-stealth, red!60, thick},
            upconv/.style={-stealth, green!60!black, thick},
            copy/.style={-stealth, gray!50, line width=0.1cm},
            copynoarrow/.style={gray!50, line width=0.1cm},
            label/.style={font=\scriptsize, align=center},
            legend/.style={font=\footnotesize, align=left}
        ]

        % Parameters of boxes
        \def\boxwidth{0.1cm}
        \def\boxheight{3cm}
        \def\scale{0.5}

        % parameters of labels
        \def\bufflabel{4}

        % Spacing between boxes
        \def\spacing{0.4cm}
        \def\vspacing{0.7cm}

        % Buffer for arrow
        \def\buffarrow{0.05cm}

        % Contracting Path (Encoder)
        \node[box={\boxheight}{\boxwidth}] (l1) at (-\spacing,0) {};
        \node[label, rotate=90] at (-\boxwidth/2-1.5*\bufflabel-\spacing, 0) {$64^2\times 1$};

        \node[box={\boxheight}{\boxwidth}] (l2) at (\boxwidth+\spacing,0) {};

        \node[box={\boxheight}{\boxwidth}] (l3) at (2*\boxwidth+2*\spacing,0) {};

        \node[box={\boxheight*\scale}{2*\boxwidth}] (l4) at (2*\boxwidth+2*\spacing,-\boxheight/2 -\boxheight*\scale/2-\vspacing) {};
        \node[label, rotate=90] at ($(l4)+(-\boxwidth-1.5*\bufflabel, 0)$) {$32^2\times 16$};

        \node[box={\boxheight*\scale}{2*\boxwidth}] (l5) at (4*\boxwidth+3*\spacing,-\boxheight/2 -\boxheight*\scale/2-\vspacing) {};

        \node[box={\boxheight*\scale}{2*\boxwidth}] (l6) at (6*\boxwidth+4*\spacing,-\boxheight/2 -\boxheight*\scale/2-\vspacing) {};

        \node[box={\boxheight*\scale^2}{4*\boxwidth}] (l7) at (6*\boxwidth+4*\spacing, -\boxheight/2 -\boxheight*\scale-\boxheight*\scale^2/2-2*\vspacing) {};
        \node[label] at ($(l7)+(-2*\boxwidth-5*\bufflabel, 0)$) {$16^2\times 32$};

        \node[box={\boxheight*\scale^2}{4*\boxwidth}] (l8) at (10*\boxwidth+5*\spacing,-\boxheight/2 -\boxheight*\scale-\boxheight*\scale^2/2-2*\vspacing) {};

        \node[box={\boxheight*\scale^2}{4*\boxwidth}] (l9) at (14*\boxwidth+6*\spacing,-\boxheight/2 -\boxheight*\scale-\boxheight*\scale^2/2-2*\vspacing) {};

        % Bottom Part (Bottleneck)
        \node[box={\boxheight*\scale^3}{8*\boxwidth}] (b1) at (14*\boxwidth+6*\spacing,-\boxheight/2 -\boxheight*\scale-\boxheight*\scale^2-\boxheight*\scale^3/2-3*\vspacing) {};
        \node[label] at ($(b1)+(-4*\boxwidth-5*\bufflabel, 0)$) {$8^2\times 64$};

        \node[box={\boxheight*\scale^3}{8*\boxwidth}] (b2) at (22*\boxwidth+7*\spacing,-\boxheight/2 -\boxheight*\scale-\boxheight*\scale^2-\boxheight*\scale^3/2-3*\vspacing) {};

        \node[box={\boxheight*\scale^3}{8*\boxwidth}] (b3) at (30*\boxwidth+8*\spacing,-\boxheight/2 -\boxheight*\scale-\boxheight*\scale^2-\boxheight*\scale^3/2-3*\vspacing) {};

        % Expansive Path (Decoder)
        \node[box={\boxheight*\scale^3}{8*\boxwidth}] (r1) at (30*\boxwidth+12*\spacing,-\boxheight/2 -\boxheight*\scale-\boxheight*\scale^2-\boxheight*\scale^3/2-3*\vspacing) {};

        \node[box={\boxheight*\scale^3}{8*\boxwidth}] (r2) at (38*\boxwidth+13*\spacing,-\boxheight/2 -\boxheight*\scale-\boxheight*\scale^2-\boxheight*\scale^3/2-3*\vspacing) {};

        \node[box={\boxheight*\scale^3}{8*\boxwidth}] (r3) at (46*\boxwidth+14*\spacing,-\boxheight/2 -\boxheight*\scale-\boxheight*\scale^2-\boxheight*\scale^3/2-3*\vspacing) {};

        \node[box={\boxheight*\scale^2}{4*\boxwidth}] (r4) at (46*\boxwidth+14*\spacing,-\boxheight/2 -\boxheight*\scale-\boxheight*\scale^2/2-2*\vspacing) {};
        \node[greenbox={\boxheight*\scale^2}{4*\boxwidth}] (r4-l) at (42*\boxwidth+14*\spacing,-\boxheight/2 -\boxheight*\scale-\boxheight*\scale^2/2-2*\vspacing) {};

        \node[box={\boxheight*\scale^2}{4*\boxwidth}] (r5) at (51*\boxwidth+15*\spacing,-\boxheight/2 -\boxheight*\scale-\boxheight*\scale^2/2-2*\vspacing) {};

        \node[box={\boxheight*\scale^2}{4*\boxwidth}] (r6) at (54*\boxwidth+16*\spacing,-\boxheight/2 -\boxheight*\scale-\boxheight*\scale^2/2-2*\vspacing) {};

        \node[box={\boxheight*\scale}{2*\boxwidth}] (r7) at (54*\boxwidth+16*\spacing,-\boxheight/2 -\boxheight*\scale/2-\vspacing) {};
        \node[greenbox={\boxheight*\scale}{2*\boxwidth}] (r7-l) at (52*\boxwidth+16*\spacing,-\boxheight/2 -\boxheight*\scale/2-\vspacing) {};

        \node[box={\boxheight*\scale}{2*\boxwidth}] (r8) at (56*\boxwidth+17*\spacing,-\boxheight/2 -\boxheight*\scale/2-\vspacing) {};

        \node[box={\boxheight*\scale}{2*\boxwidth}] (r9) at (58*\boxwidth+18*\spacing,-\boxheight/2 -\boxheight*\scale/2-\vspacing) {};

        % Residual block example
        \node[graybox={\boxheight}{\boxwidth}] (res1) at (26*\boxwidth+5*\spacing,0) {};
        \node[graybox={\boxheight}{\boxwidth}] (res2) at (27*\boxwidth+6*\spacing,0) {};
        \node[graybox={\boxheight}{\boxwidth}] (res3) at (28*\boxwidth+7*\spacing,0) {};
        \node[graybox={\boxheight}{\boxwidth}] (res4) at (29*\boxwidth+9*\spacing,0) {};
        \node[graybox={\boxheight}{\boxwidth}] (res5) at (30*\boxwidth+10*\spacing,0) {};
        \node[graybox={\boxheight}{\boxwidth}] (res6) at (31*\boxwidth+11*\spacing,0) {};
        \node[graybox={\boxheight}{\boxwidth}] (res7) at (32*\boxwidth+13*\spacing,0) {};
        \node[graybox={\boxheight}{\boxwidth}] (res8) at (33*\boxwidth+14*\spacing,0) {};
        \node[graybox={\boxheight}{\boxwidth}] (res9) at (34*\boxwidth+15*\spacing,0) {};
        \node[draw, line width = .7pt, rounded corners, inner sep=0.45cm, fit= (res1) (res2) (res3) (res4) (res5) (res6) (res7) (res8) (res9)] (internal) {};

        % Final Output Path
        \node[box={\boxheight}{\boxwidth}] (f1) at (58*\boxwidth+18*\spacing,0) {};
        \node[greenbox={\boxheight}{\boxwidth}] (f1-l) at (57*\boxwidth+18*\spacing,0) {};

        \node[box={\boxheight}{\boxwidth}] (f2) at (59*\boxwidth+19*\spacing,0) {};

        \node[box={\boxheight}{\boxwidth}] (f3) at (60*\boxwidth+20*\spacing,0) {};

        \node[box={\boxheight}{\boxwidth}] (f4) at (61*\boxwidth+22*\spacing,0) {};

        % Connections
        % Contracting Path
        \draw[connection] (l1) -- (l2) node[midway, above] {$\mathcal{P}$};
        \draw[conv] (l2) -- (l3);
        \draw[pool] (l3) -- (l4);
        \draw[conv] (l4) -- (l5);
        \draw[conv] (l5) -- (l6);
        \draw[pool] (l6) -- (l7);
        \draw[conv] (l7) -- (l8);
        \draw[conv] (l8) -- (l9);
        \draw[pool] (l9) -- (b1);

        % Bottom
        \draw[conv] (b1) -- (b2);
        \draw[conv] (b2) -- (b3);

        % Expansive Path
        \draw[conv] (r1) -- (r2);
        \draw[conv] (r2) -- (r3);
        \draw[upconv] (r3) -- (r4);
        \draw[inv] (r4) -- (r5);
        \draw[conv] (r5) -- (r6);
        \draw[upconv] (r6) -- (r7);
        \draw[inv] (r7) -- (r8);
        \draw[conv] (r8) -- (r9);
        \draw[upconv] (r9) -- (f1);
        \draw[inv] (f1) -- (f2);
        \draw[conv] (f2) -- (f3);
        \draw[connection] (f3) -- (f4) node[midway, above] {$\mathcal{Q}$};

        % Copy Connections
        \draw[copynoarrow] ($(l3) + (\boxwidth/2+\buffarrow, 0)$) -- ($(res1) - (0.5, 0)$);
        \draw[copy] ($(res9) + (0.5, 0)$) -- ($(f1-l) - (\boxwidth/2+\buffarrow, 0)$);
        \draw[copy] ($(l6) + (\boxwidth+\buffarrow, 0)$) -- ($(r7-l) - (\boxwidth+\buffarrow, 0)$);
        \draw[copy] ($(l9) + (2*\boxwidth+\buffarrow, 0)$) -- ($(r4-l) - (2*\boxwidth+\buffarrow, 0)$);
        \draw[copy] ($(b3) + (4*\boxwidth+\buffarrow, 0)$) -- ($(r1) - (4*\boxwidth+\buffarrow, 0)$);

        % Residual Connections
        \draw[connection] ($(res1) - (0.5, 0)$)-- (res1);
        \draw[conv] (res1) -- (res2);
        \draw[conv] (res2) -- (res3);
        % \draw[connection] ($(res1) + (0, \boxheight/2+1)$) to[out=70, in=110] ($(res3) + (0, \boxheight/2+1)$);
        \draw[connection] ($(res1) + (-0.25, 0)$) .. controls +(-0.5,2.4) and +(0.5,2.4) .. ($(res3) + (0.25, 0)$);
        \draw[connection] (res3) -- (res4);
        \draw[conv] (res4) -- (res5);
        \draw[conv] (res5) -- (res6);
        % \draw[connection] ($(res4) + (0, \boxheight/2+1)$) to[out=70, in=110] ($(res6) + (0, \boxheight/2+1)$);
        \draw[connection] ($(res4) + (-0.25, 0)$) .. controls +(-0.5,2.4) and +(0.5,2.4) .. ($(res6) + (0.25, 0)$);
        \draw[connection] (res6) -- (res7);
        \draw[conv] (res7) -- (res8);
        \draw[conv] (res8) -- (res9);
        % \draw[connection] ($(res7) + (0, \boxheight/2+1)$) to[out=70, in=110] ($(res9) + (0, \boxheight/2+1)$);
        \draw[connection] ($(res7) + (-0.25, 0)$) .. controls +(-0.5,2.4) and +(0.5,2.4) .. ($(res9) + (0.25, 0)$);
        \draw[connection] (res9) -- ($(res9) + (0.5, 0)$);

        % Legend
        \begin{scope}[shift={(58*\boxwidth+17*\spacing,-\boxheight/2 -\boxheight*\scale-\boxheight*\scale^2/2-2*\vspacing)}]
            \draw[conv] (0,0) -- (1,0) node[legend, right] {Convolutional} node[legend, black, yshift=10, xshift=-2] {Blocks legend:};
            \draw[copy] (0,-0.5) -- (1,-0.5) node[legend, right] {Residual};
            \draw[pool] (0,-1) -- (1,-1) node[legend, right] {Downsampling};
            \draw[upconv] (0,-1.5) -- (1,-1.5) node[legend, right] {Upsampling};
            \draw[inv] (0,-2) -- (1,-2) node[legend, right] {Invariant};
        \end{scope}
    \end{tikzpicture}
    \caption{Visual representation of a Convolutional Neural Operator with channel multiplier equal to $16$ and initial resolution equal to $64$.}
    \label{fig:cno}
\end{figure}

To resume the input function is lifted through the lifting operator, then the lifted function passes through a sequence of encoders (where the resolution is reduced and the channel size is increased), then passes through a sequence of decoders (where the resolution is increased and the channel size is decreased) and finally the output function is obtained through the projection operator. More precisely, let $r$ be the initial resolution and $c$ be the initial number of channels relative to the initial data. We define a hyperparameter called the channel multiplier ($chan\_mul$) such that the CNO architecture is constructed to have, at the $i$-th layer for $i = 1, \dots, L$, a resolution equal to $r/2^{i-1}$ and a number of channels equal to $c$ in the first (input) layer and $ 2^{i-1}\cdot c \cdot chan\_mul$, $i\ge 2$. For example, in Figure \ref{fig:cno} we have $r=64$, $c=1$ and $chan\_mul = 16$. Furthermore, encoders and decoders are connected through skip connections using the residual and invariant operators block, this is needed for the transfer of high-frequency content to the output before filtering them out with the downsampling block. For a visual representation of the CNO architecture refer to Figure \ref{fig:cno}. The figure illustrates the flow of the input function through the lifting operator, encoder-decoder sequence, and projection operator, as well as the skip connections and residual blocks that connect the encoders and decoders with different arrows and colours.

\subsubsection*{Convolution operator}
The convolution operator (or convolution block) $\mathcal{K}_{\omega, \theta}: \mathcal{B}_{\omega}(D, \R^{d_v}) \to \mathcal{B}_{\omega}(D, \R^{d_v})$ is the analogue of the kernel integral operator for the FNOs, where the suffix $\omega$ indicates that the operator maps functions of bandwidth $\omega$ and the suffix $\theta$ indicates that the operator is parameterized by some learnable parameters. We make the following assumptions
\[
    \mathcal{K}_{\omega, \theta}f(x) = (f * \kappa_{\omega, \theta})(x) = \int_D \kappa_{\omega, \theta}(x)f(x-y)dy,  \quad f \in \mathcal{B}_{\omega}(D, \R^{d_v}), \quad \forall x \in D.
\]
In the previous formula  $\kappa_{\omega, \theta}(x) \in \R^{d_v \times d_v}$, usually called kernel function, and $f(x-y) \in \R^{d_v}$ for every $x, y \in D$. Therefore discretizing and truncating the integral, we obtain
\begin{equation}
    \mathcal{K}_{\omega, \theta}f(x) = \int_D \kappa_{\omega, \theta}(y)f(x-y)dy \approx \sum_{i,j=1}^k \kappa_{\omega, \theta}(y_{ij})f(x - y_{ij}) = \sum_{i,j=1}^k \kappa_{ij} \cdot f(x - y_{ij}),  \quad \forall x \in D.
    \label{eq:convolution_operator_cno}
\end{equation}
In this context, $k$ is usually called the kernel size of the convolution and $d_v$ is called the channel dimension.
\\
Following the definition of the convolution operator in \eqref{eq:convolution_operator_cno}, we can see that the kernel function is parameterized in physical space instead of Fourier space as in \eqref{eq:discretized_integral_operator_fno}. So our parameterization is of a \textit{local} nature, but we recover non-locality by composing several of these layers.

\subsubsection*{Upsampling and downsampling operators}
For some $\overline{\omega} > \omega$, we can upsample a function $f \in \mathcal{B}_{\omega}(D, \R)$ to the higher band $\mathcal{B}_{\overline{\omega}}(D, \R)$ by simply setting,
\[\mathcal{U}_{\omega,\overline{\omega}} : \mathcal{B}_{\omega}(D, \R) \to \mathcal{B}_{\overline{\omega}}(D, \R), \quad \mathcal{U}_{\omega,\overline{\omega}}f(x) = f(x), \quad \forall x \in D.\]
\noindent
This is trivial since $f \in \mathcal{B}_{\omega}(D, \R) \subset \mathcal{B}_{\overline{\omega}}(D, \R)$, so upsampling does not change the function itself; it simply reinterprets $f$ as belonging to the higher bandlimited space.

On the other hand, for some $\underline{\omega} < \omega$, we can downsample a function $f \in \mathcal{B}_{\omega}(D, \R)$ to the lower band $\mathcal{B}_{\underline{\omega}}(D, \R)$ by setting $\mathcal{D}_{\omega,\underline{\omega}} : \mathcal{B}_{\omega}(D, \R) \to \mathcal{B}_{\underline{\omega}}(D, \R)$, defined by
\[\mathcal{D}_{\omega,\underline{\omega}}f(x) = \left(\frac{\underline{\omega}}{\omega}\right)^2 (h_{\underline{\omega}} *f)(x) = \left(\frac{\underline{\omega}}{\omega}\right)^2 \int_D h_{\underline{\omega}}(x-y)f(y)dy, \quad \forall x \in D,\]
\noindent
where $\left(\frac{\underline{\omega}}{\omega}\right)^2$ is a proper normalization factor and $h_{\underline{\omega}}$ is the so-called interpolation sinc filter, that is a low-pass filter of $f$ that removes frequencies higher than $\underline{\omega}$, defines as:
\[h_{\omega}(x) = \prod_{i=1}^{d}\text{sinc}(2\omega x_i) , \quad x \in \mathbb{R}^d,\]
where $\text{sinc}(x) = \sin(\pi x)/(\pi x)$ is the sinc function.

For function that are vector-valued $f \in \mathcal{B}_{\omega}(D, \R^d)$, for some $d \in \N$ and $d>1$, we define the upsampling operator $\mathcal{U}_{\omega,\overline{\omega}}: \mathcal{B}_{\omega}(D, \R^d) \to \mathcal{B}_{\overline{\omega}}(D, \R^d)$ and the downsampling operator $\mathcal{D}_{\omega,\underline{\omega}}: \mathcal{B}_{\omega}(D, \R^d) \to \mathcal{B}_{\underline{\omega}}(D, \R^d)$ componentwise, i.e. we apply the scalar definition to each dimension independently.

\subsubsection*{Activation operator}
Naively, it is common in machine learning to apply the activation function pointwise to any function. However, it has been shown (see \cite{reno24bar}) that such an application can generate features that do not respect the bandlimits of the underlying function space, leading to aliasing errors. To avoid aliasing errors and satisfy the CDE condition, we first upsample the input function $f \in \mathcal{B}_\omega(D, \R)$ to a higher bandlimit $\overline{\omega} > \omega$, then apply the activation, assuming that $\overline{\omega}$ is large enough so that $\sigma(\mathcal{B}_\omega(D, \R)) \subset \mathcal{B}_{\overline{\omega}}(D, \R)$, and finally downsampling the result back to the original band limit $\omega$. So we define the activation operator, or convolutional block, as
\[\Sigma_{\omega} : \mathcal{B}_\omega(D, \R) \to \mathcal{B}_\omega(D, \R), \quad \Sigma_{\omega}f(x) = (\mathcal{D}_{\overline{\omega},\omega}\circ \sigma \circ \mathcal{U}_{\omega,\overline{\omega}}f)(x), \quad \forall x \in D.\]

As with the upsampling and downsampling operators, we define the activation operator for the vector-valued function $\mathcal{B}_{\omega}(D, \R^{d})$ componentwise.

\subsubsection*{Residual and invariant operators}
The residual and invariant operators are defined in relation to the previously defined operators. The residual (or skip-connection) operator connects the encoder and decoder blocks with the same spatial resolution, and this is useful for transferring high frequency content to the output before it is filtered out by the downsampling block. In particular, the output of the residual block is concatenated with the corresponding decoder block. The residual operator is a concatenation of the convolutional operator and the activation operator with skip connections; the architecture emulates the classical Residual Neural Network, in particular
\[
    \mathcal{R}_{\omega, \theta} : \mathcal{B}_{\omega}(D, \R^{d_v}) \to \mathcal{B}_\omega(D, \R^{d_v}), \quad  \mathcal{R}_{\omega, \theta}(v) = v + \mathcal{K}_{\omega, \theta} \circ \Sigma_{\omega} \circ \mathcal{K}_{\omega, \theta}(v), \quad \forall v \in \mathcal{B}_\omega(D, \R^{d_v}).
\]
Where we indicate with $D \subset \R^d$ the domain and $d_v \in \N$ the number of channels of the input function.

At every level of the architecture, after the concatenation between the output of the residual block and the output of the decoder block, we apply the invariant block that is defined as
\[
    \mathcal{I}_{\omega, \theta} : \mathcal{B}_\omega(D, \R^{d_v}) \to \mathcal{B}_\omega(D, \R^{d_v}), \quad  \mathcal{I}_{\omega, \theta}(v) = \Sigma_{\omega} \circ \mathcal{K}_{\omega, \theta}(v), \quad \forall v \in \mathcal{B}_\omega(D,\R^{d_v}).
\]

\subsection{Comparison of FNOs and CNOs}
While both FNOs and CNOs are powerful architectures for learning mappings between function spaces, they differ in their mathematical foundations:
\begin{itemize}
    \item \textbf{Global vs. local discretization:} FNOs discretize the kernel function in Fourier space and thus use a global discretization at each layer; instead, CNOs discretize the kernel function in physical space using a local discretization at each layer.

    \item \textbf{Computational efficiency:} FNOs use the Fast Fourier Transform (FFT) for efficient computation, while CNOs rely on generalised convolution operations, which can be computationally expensive for large domains.

    \item \textbf{Resolution independence:} Both FNOs and CNOs are resolution independent, meaning that they can handle input functions of different sizes without retraining. Only CNOs have the property of being representative equivalent neural operators and satisfying the CDE condition, which is a fundamental theoretical property for neural operators.

    \item \textbf{Translation equivariance:} CNOs inherit the translation equivalence property from the convolution operator, while FNOs do not. This property translates into better generalisation capabilities of the model when the problem is expected to be invariant under translations of the input, such as the transport equation, as numerically demonstrated in \cite{CNO23raonic}.
\end{itemize}

Fourier neural operators and convolutional neural operators are two powerful approaches to learning mappings between function spaces. Together, these architectures provide a versatile toolkit for solving complex problems in scientific computing, engineering, and beyond. The development of automated hyperparameter optimization routines, as discussed in the previous section, further enhances the practicality and scalability of these methods, paving the way for their widespread adoption in real-world applications.

%%%%%%%%%%%%%%%%%%%%%%%%%%%%%%%%%%%%%%%%%
\section{Explanation of HyperNOs library design}\label{section:hypernos_model_problem}
We would like to provide a brief overview of the design of the HyperNOs library, focusing on the key components and features that make it a useful tool for studying neural operators. The library is built around a modular architecture that allows users to define custom search spaces, models, data sets and loss functions, making it highly flexible and extensible with a user-friendly interface. In this section we will explain the library using a model problem and an explicit code example.

\subsection{Model problem and dataset}
As a model problem, we consider the Darcy problem, which is a classical example of a partial differential equation that arises in many applications and has often been used as a starting point for experimentally testing the behaviour of neural operators. We consider the Darcy flow equation in two dimensions, subject to homogeneous Dirichlet boundary conditions, the second-order elliptic equation, given by the following equation
\begin{equation}
    \begin{cases}
        - \nabla(a \cdot \nabla u) = f,\quad & \mathrm{in}\ D           \\
        u = 0,                               & \mathrm{in} \ \partial D \\
        f = 1,                               & \mathrm{in} \ D
    \end{cases}
    \label{eq:darcy_model_problem}
\end{equation}
where $ D = (0, 1)^2 $ is the unit square and we fix the source term equal to $1$. In this setting $ \mathcal{A} = L^{\infty}(D, \R^+) $, $ \mathcal{U} = H^1_0(D, \R) $. We consider the weak form of the solution operator to be the form
\[  \mathcal{G}: L^{\infty}(D, \R^+) \to H^1_0(D, \R), \qquad \mathcal{G}:a \mapsto u. \]
As a dataset, we consider the same created in \cite{CNO23raonic}, where the input is the diffusion coefficient $a \sim \psi\#\mu$, with $\mu$ being a Gaussian Process with zero means and squared exponential kernel
\[
    k(x,y) = \sigma^2 \exp \left(\frac{|x-y|^2}{l^2}\right), \quad \sigma^2 = 0.1,
\]
with $l = 0.1$ for the in-distribution testing and $l = 0.05$ for the out-of-distribution testing. The mapping $\psi: \mathbb{R} \to \mathbb{R}$ takes the value $12$ on the positive part of the real line and $3$ on the negative part.

This dataset is already implemented in the library, specifically in the \texttt{datasets.py} module, and can be easily loaded and used to train different neural operators. The dataset can be downloaded by simply executing the file \texttt{download\_data.sh} present in the library, and it can be loaded using the \texttt{Darcy} class, which takes as input the number of Fourier features we want to include, the seed number (if the seed is positive, it is fixed at this number for reproducibility, otherwise no seed is fixed), the dimension of the batch size, and the number of training samples. You can see the code for loading the dataset in Code \ref{code:model_problem} from line $44$ to line $48$.

There are other datasets already implemented in the library, in particular, we have implemented all the functionalities useful for the datasets from \cite{CNO23raonic}. All the other functions implemented to load the provided datasets can be used in a similar way to the one presented above. Of course, the user can implement his own dataset by inheriting from the \texttt{torch.utils.data.Dataset} class. The requirements for the dataset are to have the batch dimension as the first dimension and to be divided into three subsets: the training set for the training process, the validation set for the hyperparameter optimization process and the test set for final testing.

\subsection{Model architecture}
The model architecture is defined by the user via a model builder function, which takes a configuration dictionary as input and returns an instance of the neural operator model. The model builder function is passed to the \texttt{tune\_hyperparameters} function, which is the main function of the library and is responsible for handling the hyperparameter optimization process. The configuration dictionary contains all the necessary hyperparameters that define the neural operator $\mathcal{G}_{\theta}$. For this particular example, we consider the architecture of the Fourier neural operator implemented in the \texttt{FNO.py} module of the library, we implement it according to the Section \ref{subsection:fno} and try to be as general as possible. Within the library, we have even implemented the latest architecture of convolutional neural operators in the \texttt{CNO.py} module, following Section \ref{subsection:cno} and trying to implement as many features as possible. The user can define his own model architecture by implementing his own class of the model and the associated custom model builder function to be passed to \texttt{tune\_hyperparameters}. The only requirement is that the model must take only one input in the forward method, and the tensors passed as input to the model need to have the as first dimension the batch dimension then the space dimension and the last dimension is the input dimension $(d_a)$. The code for the model builder function is shown in Code \ref{code:model_problem} from line $35$ to line $42$.

\subsection{Loss function}
The loss function is defined by the user via a loss function object which is passed to the \texttt{tune\_hyperparameters} function. The loss function object should implement the \texttt{\_\_call\_\_} method, which takes as input the model output and the target values (in that order) and returns the sum of the loss value along the batch dimension. In this example we consider the $L^1$ relative loss function discussed in \ref{subsection:data_discretization} and implemented in the \texttt{loss\_fun.py} module of the library. The code for the loss function object is shown in Code \ref{code:model_problem} on line $50$.

The user can find already implemented the $L^p(\Omega), \ \forall p \in [1, +\infty)$ and for all the domains $\Omega$, and $H^p(D), \ \forall p \in [1, +\infty) $ for rectangular domains $D$. The user can define their own loss function by implementing a custom loss function object that satisfies the requirements described above.

\subsection{Hyperparameter optimization}
The last necessary step is to define the hyperparameter search space and call the \texttt{tune\_hyperparameters} function, which handles the hyperparameter optimization process. The search space is defined by a configuration dictionary, where each key corresponds to a hyperparameter and the value is a tune search space object. The user can define the search space using the \texttt{tune} module from the Ray library \cite{ray18moritz}, which provides a wide range of search space objects, including \texttt{choice}, \texttt{randint}, \texttt{uniform}, \texttt{loguniform} and \texttt{randn}. The config space must contain all the hyperparameters that define the neural operator $\mathcal{G}_{\theta}$, the data set, the loss function, and the hyperparameters for the optimization process. The code for this example is shown in code \ref{code:model_problem} from line $10$ to line $34$. We have fix some reasonable hyperparameters, such as the number of examples in the train set and the validation set, the number of epochs, the batch size, and many others. We also decided how to vary the more interesting hyperparameters, such as the number of layers, the width of the network, the number of modes, the activation function, the architecture of the network, the learning rate, the weight decay, and so on. This search space can be easily modified by the user to find the best choice of hyperparameters for an architecture or to study the dependence of the architecture's performance on a few hyperparameters.

After defining the search space, the dataset builder, the model builder and the loss function, we can call the \texttt{tune\_hyperparameters} function, which performs the hyperparameter optimization process using the Ray-tune library, see lines $52-55$ of Code \ref{code:model_problem}. The function has other optional arguments that can be used to customize the optimization process. The number of sample parameters determines how many trials to run to find the best possible hyperparameter. We can decide how many CPU cores and GPUs to use in each trial of the optimization process, and the number of trials to run in parallel, depends on the available hardware and this is automatically calculated by the library. In our example Code \ref{code:model_problem} we decide to allocate half of each GPU to each trial, in other words, the library runs two trials in parallel for each GPU present on the machine. The optimization process is based on the HyperOpt Python library, which mainly uses the Tree-structured Parzen Estimator (TPE) approach and the adaptive TPE, see \cite{hyperopt11bergstra}. The library also uses the ASHA algorithm to stop the bad trials to speed up the overall process, see \cite{asha20li}. Finally, the library automatically saves checkpoints of the trials, which can be resumed at any time, and there is Tensorboard support for better visualization of the results.

\subsection{Final training}
After finding the best hyperparameters for the model, the user can train the model using the best hyperparameters found in the optimization process. The user can use the supplied \texttt{train\_fixed\_model} function, which takes as input the chosen hyperparameters, the model builder, the dataset builder and the loss function, just as before. We also need to provide a name to identify the test, a function to plot the input and a function to plot the output. The function trains the model on the test set and saves the trained model with the chosen name. During training, the loss function values are stored togheter with the outputs and the error of the model using Tensorboard for better visualisation of the training process.

Practically, the code for the final testing is similar to the one presented in the code \ref{code:model_problem}, the only difference being that we don't have the config space since every hyperparameter is fixed by the user. The data set builder, model builder and loss function are the same as before and the call to the function \texttt{train\_fixed\_model} is the only new thing the user has to do. The code for the final training is shown in Code \ref{code:final_training}.

\begin{code}[!ht]
    \begin{pythoncode}{{\color{black}{\textbf{Final training}}}}
train_fixed_model(
    default_hyper_params, # fixed hyperparameters
    model_builder, # model builder function, defined as before
    dataset_builder, # dataset builder function, defined as before
    loss_fn, # loss function object, defined as before
    experiment_name, # name of the experiment
    get_plot_function("poison", "input"), # function to plot the input
    get_plot_function("poison", "output"), # function to plot the output
)
    \end{pythoncode}
    \caption{Code for the final training.}
    \label{code:final_training}
\end{code}

\newpage
\subsection{Code for the model problem}
% !TeX ignore = begin
\begin{code}[!ht]
    \begin{pythoncode}{{\color{black}{\textbf{Model problem}}}}
import torch
from datasets import Darcy
from FNO.FNO import FNO
from loss_fun import LprelLoss
from ray import tune
from tune import tune_hyperparameters

device = torch.device("cuda" if torch.cuda.is_available() else "cpu")

config_space = {
    "FourierF": tune.choice([0]),
    "RNN": tune.choice([False]),
    "batch_size": tune.choice([32]),
    "epochs": tune.choice([1000]),
    "fft_norm": tune.choice([None]),
    "fno_arc": tune.choice(["Classic", "Zongyi", "Residual"]),
    "fun_act": tune.choice(["tanh", "relu", "gelu", "leaky_relu"]),
    "in_dim": tune.choice([1]),
    "include_grid": tune.choice([1]),
    "learning_rate": tune.quniform(1e-4, 1e-2, 1e-5),
    "modes": tune.choice([2, 4, 8, 12, 16, 20, 24, 28, 32]),
    "n_layers": tune.randint(1, 6),
    "out_dim": tune.choice([1]),
    "padding": tune.randint(0, 16),
    "problem_dim": tune.choice([2]),
    "retrain": tune.choice([4]),
    "scheduler_gamma": tune.quniform(0.75, 0.99, 0.01),
    "scheduler_step": tune.choice([10]),
    "training_samples": tune.choice([256]),
    "val_samples": tune.choice([128]),
    "weight_decay": tune.quniform(1e-6, 1e-3, 1e-6),
    "weights_norm": tune.choice(["Kaiming"]),
    "width": tune.choice([4, 8, 16, 32, 64, 128, 256])}

model_builder = lambda config: FNO(
    config["problem_dim"],
    config["in_dim"],config["width"],
    config["out_dim"], config["n_layers"],
    config["modes"], config["fun_act"],
    config["weights_norm"], config["fno_arc"],
    config["RNN"],config["fft_norm"],
    config["padding"], device, config["retrain"])

dataset_builder = lambda config: Darcy(
    {"FourierF": config["FourierF"],
     "retrain": config["retrain"]},
    batch_size=config["batch_size"],
    training_samples=config["training_samples"])

loss_fn = LprelLoss(1)

tune_hyperparameters(
    config_space, model_builder,
    dataset_builder, loss_fn, num_samples=50,
    runs_per_cpu=0.0, runs_per_gpu=1.0)

    \end{pythoncode}
    \caption{Code for the model problem.}
    \label{code:model_problem}
\end{code}
% !TeX ignore = end
\newpage
%%%%%%%%%%%%%%%%%%%%%%%%%%%%%%%%%%%%%%%%%
\section{Advanced features with examples}\label{section:advanced_examples}
In this section, we explain many different modifications of the basic Code \ref{code:model_problem} that can be used to solve more complex and interesting problems involving neural operators. The features presented are completely independent, so they can be combined with each other.

\subsection{Default hyperparameters loading}\label{subsection:default_hyperparams}
Within the library we provide the hyperparameters that have been identified as optimal for a variety of datasets, these hyperparameters have been carefully tuned through extensive experimentation, as detailed in Section \ref{section:numerical_experiments}. Making these configurations readily available ensures reproducibility, detailed debugging, efficiency and scalability in experiments. This approach allows researchers to quickly replicate results, analyse model behaviour, and build on existing work without redundant hyperparameter tuning, focusing on higher-level tasks such as model interpretation and application.  For readability reasons, the hyperparameters are stored in json files and divided into two dictionaries: those needed for the training process and those needed for the optimization process. This step ensures that all necessary configurations are available in a consistent format, simplifying the setup process. This structured approach improves the user experience, code readability and promotes best practices in machine learning experimentation. With this in mind, we have implemented some functionality for the user to load and use the desired hyperparameters. An example of how to load and merge these hyperparameters is provided in Code \ref{code:load_default_hyperparameters}.

\begin{code}[!ht]
    \begin{pythoncode}{{\color{black}{\textbf{Load default hyperparameters}}}}
# Load the best hyperparameters for the FNO model of the Poisson example
hyperparams_train, hyperparams_arc = FNO_initialize_hyperparameters(
    which_example="poisson", mode="best")

# Alternatively, load the best hyperparameters for the CNO model
hyperparams_train, hyperparams_arc = CNO_initialize_hyperparameters(
    which_example="poisson", mode="best")

# Merge the hyperparameters togheter
fixed_params = { **hyperparams_train, **hyperparams_arc}
    \end{pythoncode}
    \caption{Code for loading the default hyperparameters.}
    \label{code:load_default_hyperparameters}
\end{code}

In addition to the advantages already discussed, the default hyperparameters can also be used within the hyperparameter optimization process itself. When using the \texttt{tune\_hyperparameters} function, the default hyperparameters can be passed as the initial configuration. This allows the optimization process to start by evaluating the default hyperparameters, which serve as a baseline for comparison with other trials. By using the ASHA algorithm, the optimization process can quickly eliminate poorly performing trials and focus computational resources on more promising configurations. This significantly speeds up the overall optimization process by allowing the algorithm to prioritize trials that are likely to yield better results.
\\
In addition, this approach allows users to focus on optimizing only the most critical parameters, while leaving the others at their default values. This eliminates the need to explicitly specify all hyperparameters, reducing the complexity of the configuration space and speeding up the optimization process. For example, if a user is primarily interested in optimizing the learning rate and number of layers, they can leave the remaining hyperparameters (e.g. activation function, batch size, etc.) at their default values.
\\
The modifications required to implement this feature in the configuration space and in the optional argument of the \texttt{tune\_hyperparameters} function are demonstrated in Code \ref{code:default_hyperparameters}. This implementation ensures that default hyperparameters are seamlessly integrated into the optimization workflow, providing a robust and efficient framework for hyperparameter tuning.

\begin{code}[!ht]
    \begin{pythoncode}{{\color{black}{\textbf{Load default hyperparameters for optimization}}}}
hyperparams_train, hyperparams_arc = FNO_initialize_hyperparameters(
    "poison", mode=mode_hyperparams)

config_space = {
    "fno_arc": tune.choice(["Classic", "Zongyi", "Residual"]),
    "fun_act": tune.choice(["tanh", "relu", "gelu", "leaky_relu"]),
    "learning_rate": tune.quniform(1e-4, 1e-2, 1e-5),
    "modes": tune.choice([2, 4, 8, 12, 16, 20, 24, 28, 32]),
    "n_layers": tune.randint(1, 6),
    "padding": tune.randint(0, 16),
    "scheduler_gamma": tune.quniform(0.75, 0.99, 0.01),
    "weight_decay": tune.quniform(1e-6, 1e-3, 1e-6),
    "width": tune.choice([4, 8, 16, 32, 64, 128, 256])}

# Set all the other parameters to fixed values
fixed_params = {**hyperparams_train, **hyperparams_arc}
parameters_to_tune = config_space.keys()
for param in parameters_to_tune:
    fixed_params.pop(param, None)
config_space.update(fixed_params)

tune_hyperparameters(
    config_space, model_builder, dataset_builder,
    loss_fn, num_samples=50,
    default_hyper_params=[{**hyperparams_train, **hyperparams_arc}]
    runs_per_cpu=0.0, runs_per_gpu=1.0)
    \end{pythoncode}
    \caption{Code for using the default hyperparameters in the hyperparameters optimization process.}
    \label{code:default_hyperparameters}
\end{code}

\subsection{Model with the same number of trainable parameters}\label{subsection:same_number_of_params}
In some cases, it may be desirable to compare models with the same number of trainable parameters to evaluate their performance under similar complexity constraints. To facilitate this comparison, we provide a utility function that allows the user to specify the desired number of trainable parameters and the library will automatically adjust the model architecture accordingly. To achieve this, the function calculates the number of trainable parameters for a default model architecture and then generates only trials with models with a similar number of parameters. Alternatively, the user can directly enter a number of trainable parameters and the library will generate only architectures with approximately those values of trainable parameters. This feature is interesting in application and indeed allows users to make fair comparisons between models with different hyperparameters, ensuring that performance differences are not solely due to differences in complexity. Moreover, this approach can be interesting for a more intensive study of the hyperparameter space, since the hyperparameter space is limited by the addition of this additional constraint.

We need to calculate the number of trainable parameters of the model, given the hyperparameter configuration. For the Fourier neural operator architecture, the number of trainable parameters can be approximated as follows
\begin{equation}
    \text{trainable parameters} \approx 2^{d}\cdot L \cdot d_v^2 \cdot modes^d,
    \label{eq:trainable_parameters_fno}
\end{equation}
where $d$ is the problem domain dimension, $L$ is the number of layers, $d_v$ is the hidden dimension, and $modes$ is the number of Fourier modes considered. Thus, in order to generate trials with the same number of trainable parameters, we carry out the optimization on the number of layers and the hidden dimensions; once they are chosen, we compute the number of Fourier modes according to the formula \eqref{eq:integral_operator_fno} to match as closely as possible the number of parameters of the standard architecture. In addition, we have to impose that the number of modes be an integer with positive values and satisfy the Nyquist-Shannon sampling theorem \cite{vetterli14signal}. The modification required to implement this feature is simply to remove the \textit{modes} argument from the configuration space, determine the number of modes and then pass it directly to the model builder, for clarity this is done with helper functions but the implementation is straightforward. The main changes are shown in Code \ref{code:same_trainable_parameters_fno}.

\begin{code}[!ht]
    \begin{pythoncode}{{\color{black}{\textbf{Same number of trainable parameters for FNOs}}}}
from FNO.FNO_utilities import FNO_initialize_hyperparameters, count_params_fno, compute_modes

# Load the baseline model and compute the total of params
hyperparams_train, hyperparams_arc = FNO_initialize_hyperparameters(
    which_example="poisson", mode="default")
total_default_params = count_params_fno(fixed_params, accurate=False)

config_space = {...} # Define as before, but without the modes

# Compute the correct number of modes
maximum = 32
modes = compute_modes(total_default_params, maximum, config),

model_builder = lambda config: FNO(...) # Define as before with the computed modes
    \end{pythoncode}
    \caption{Code for the model with the same number of trainable parameters.}
    \label{code:same_trainable_parameters_fno}
\end{code}

We can do the same with the convolutional neural operator architecture, the number of trainable parameters can be well approximated by the following formula:
\begin{equation}
    \text{trainable parameters} \approx k^d \cdot chan\_mul^2 \left( 2^{2L-1}M + 2N\left(\frac14 + \frac{4^{L-1}-1}{3}\right) + \frac{31}{6} 4^{L-1} - \frac{11}{12} \right),
    \label{eq:trainable_parameters_cno}
\end{equation}
where $k$ is the kernel size of the convolution, $chan\_mul$ is the value of the channel multiplier, $L$ is the number of decoding/encoding layers, $M$ is the number of layers in the last remaining block, $N$ is the number of layers in all other remaining blocks, and $d$ is the problem domain dimension. Obtaining the approximation \eqref{eq:trainable_parameters_cno} of the number of parameters is less trivial with respect to FNO, but the modifications needed to implement this feature are minimal, see Code \ref{code:same_trainable_parameters_cno}. In particular, we optimize $k, L, M, N$ (or a subset of them) and compute $chan\_mul$ to generate trials approximating the same number of hyperparameters.

\begin{code}[!ht]
    \begin{pythoncode}{{\color{black}{\textbf{Same number of trainable parameters for CNOs}}}}
from CNO.CNO_utilities import CNO_initialize_hyperparameters, count_params_cno, compute_channel_multiplier

# Load the baseline model and compute the total of params
hyperparams_train, hyperparams_arc = CNO_initialize_hyperparameters(
    which_example="poisson", mode="default")
total_default_params = count_params_cno(fixed_params, accurate=False)

config_space = {...} # Define as before, but without channel multiplier

# Compute the correct number of modes
channel_multiplier=compute_channel_multiplier(total_default_params, config)

model_builder = lambda config: CNO(...) # Define as before with the computed channel multiplier
    \end{pythoncode}
    \caption{Code for the model with the same number of trainable parameters.}
    \label{code:same_trainable_parameters_cno}
\end{code}

\subsection{Multiple datasets concatenation}\label{subsection:multiple_datasets}
In many real-world applications, it is common to work with multiple datasets that contain complementary information or represent different aspects of the same problem. In scientific computing, for example, researchers may need to combine data from different experiments, simulations or sources to build a more comprehensive model. To address this need, our library provides support for concatenating multiple data sets into a single training set, allowing users to leverage diverse data sources for model training. This feature allows researchers to incorporate a wide range of data types, formats and sources into their models, improving model robustness, generalisation and performance. Since the architecture definition in the neural operator settings does not depend on the discretization resolution of the dataset, the library allows the concatenation of datasets with different resolutions.

The code for concatenating multiple datasets is shown in Code \ref{code:multiple_datasets}. Here the \texttt{NO\_load\_data\_model} function is just a wrapper that associates the dataset name with the dataset class, and the \texttt{concat\_datasets} function concatenates the datasets into a single set. The user can specify which datasets to include in the concatenation by providing a list of dataset names, which we provide, or use custom datasets by including them in the \texttt{NO\_load\_data\_model} wrapper, or use them directly as in Code \ref{code:model_problem}.

\begin{code}
    \begin{pythoncode}{{\color{black}{\textbf{Multiple dataset concatenation}}}}
from datasets import NO_load_data_model, concat_datasets
which_example = ["poisson", "darcy"]
dataset_builder = lambda config: concat_datasets(
    *(NO_load_data_model(
    dataset_name,
    no_architecture={"FourierF": config["FourierF"],
            "retrain": config["retrain"]},
    batch_size=config["batch_size"],
    training_samples=config["training_samples"])
    for dataset_name in which_example))
    \end{pythoncode}
    \caption{Code for concatenating multiple datasets.}
    \label{code:multiple_datasets}
\end{code}

\subsection{Multiple resolution datasets training}
We have implemented the ability to train the model on the same dataset but loaded with different resolutions. This feature is useful when the user wants to train the model on the same dataset but with different resolutions, for example, to study the effect on model performance or model generality. This flexibility allows users to combine datasets with different levels of detail, granularity and complexity, providing a technique for data augmentation with a more comprehensive and accurate representation of the underlying problem. We have already mentioned all the necessary ingredients in Section \ref{subsection:multiple_datasets}, with the difference that this time we do not vary the name of the dataset, but the resolution of the discretization. We just have to be sure that the downsampled resolution must divide the original resolution. Furthermore, for the FNO architecture, the minimum resolution must be compatible with the maximum number of modes via the Nyquist-Shannon theorem. The code for this function is shown in Code \ref{code:multiple_resolution_datasets}.

\begin{code}
    \begin{pythoncode}{{\color{black}{\textbf{Multiple dataset resolutions}}}}
from datasets import NO_load_data_model, concat_datasets
resolution = [64, 32, 16] # max number of modes is 9
dataset_builder = lambda config: concat_datasets(
    *(NO_load_data_model(
    which_example="poisson",
    no_architecture={"FourierF": config["FourierF"],
            "retrain": config["retrain"]},
    batch_size=config["batch_size"],
    training_samples=config["training_samples"],
    in_size=res)
    for res in resolution))
    \end{pythoncode}
    \caption{Code for concatenating multiple datasets.}
    \label{code:multiple_resolution_datasets}
\end{code}

\subsection{Physics informed neural operators}

The architecture of the neural operators ensures a continuous representation with respect to the space variables. These features make it natural to introduce a priori physical knowledge into the construction and training of the architecture. We distinguish between weak and strong imposition. The strong imposition of physical knowledge is discussed in the next subsection. Instead, weak imposition consists of introducing physically informed terms into the loss function in order to promote solutions that satisfy certain physical requirements \cite{pino24li, pdon21wang}. The loss function is usually modified as follows
\[ \mathcal{L}oss = \mathcal{L}oss_{data} + \alpha_{phys}\mathcal{L}oss_{phys},\]
where $\mathcal{L}oss_{data}$ is the data-driven loss function defined in Section \ref{subsection:data_discretization}, $\mathcal{L}oss_{phys}$ is the loss function derived from the PDE that define the physics of the problem and $\alpha_{phys}\in \R$ is a constant scaling that can be considered as another hyperparameter to add at the tuning process. A typical example of a physics-based loss function is the norm of the residual of the PDE (strong or weak formulation) that the neural operators have to approximate. This process is fundamental to Physically Informed Neural Networks (PINNs, \cite{pinn19raissi}) and Physics Informed Neural Operator (PINO, \cite{pino24li}), but it is shown that this additional term can complicate the training procedure \cite{physloss24deryck}. The modification of the loss function to include physically informed loss is in Code \ref{code:physics_informed_loss}.

\begin{code}
    \begin{pythoncode}{{\color{black}{\textbf{Physics-informed loss function}}}}
from loss_fun import loss_selector
loss_fn = LprelLoss(2)

from loss_fun_with_physics import PoissonResidualFiniteDiff
alpha_phys = 0.01
loss_phys = PoissonResidualFiniteDiff(alpha=alpha_phys, p = 2) # norm 2 of the residual

tune_hyperparameters(
    config_space, model_builder,
    dataset_builder, loss_fn,
    default_hyper_params,
    loss_phys = loss_phys)
    \end{pythoncode}
    \caption{Code for defining a physics-informed loss function.}
    \label{code:physics_informed_loss}
\end{code}

\subsection{Model's wrapper for complex tasks}
It can be useful to define some post-processing or some modification of the architecture of the neural operator to ensure some properties or physical constraints. This is done by the model wrapper, a wrapper is a class with an initialization method that defines the model and a forward method that takes the input from the network, evaluates the model and then modifies the output before returning it.

For example, we can use wrappers to impose strong physical knowledge. If we know that on a given part of the boundary, the solution must satisfy a fixed Dirichlet boundary condition $u(x) = g(x)$ in $\Gamma_D \subset \partial D$, then we can modify the output of the model to automatically satisfy the constraint. In particular, we define
\[ u(x) = \psi(x) \mathcal{G}_{\theta}a(x) + g(x), \]
where $\psi(x)=0$ in $\Gamma_D$ and $g(x)$ is the Dirichlet boundary condition extended to zero outside of $\Gamma_D$.

Another useful case where wrappers are needed is in the definition of the Fourier continuation for FNOs \cite{uniFNO21kov}. If we have a non-rectangular domain, we can embed it in a rectangular one and then use FNOs to approximate the padded operator. When we need to compute the loss function, we want to consider only the points on the original (unpadded) domain, and this can be done with a simple mask using wrappers. The code needed to do this is shown in Code \ref{code:wrapper_fno_continuation}.

\begin{code}
    \begin{pythoncode}{{\color{black}{\textbf{Fourier continuation wrapper}}}}
# This is implemented inside the library
class AirfoilWrapper(nn.Module):
def __init__(self, model):
    super(AirfoilWrapper, self).__init__()
    self.model = model

    def forward(self, input_batch):
        output_batch = self.model(input_batch)
        output_batch[input_batch == 1] = 1
        return output_batch

    def __getattr__(self, name):
        if name != "model" and hasattr(self.model, name):
        return getattr(self.model, name)
        return super().__getattr__(name)

def wrap_model_builder(model_builder):
    def wrapped_builder(config):
        model = model_builder(config)
        return AirfoilWrapper(model)
    return wrapped_builder
    \end{pythoncode}
    \begin{pythoncode}{{\color{black}{\textbf{Fourier continuation wrapper}}}}
model_builder = lambda config: FNO(...) # Define as in Code 1
model_builder = wrap_model_builder(model_builder)
    \end{pythoncode}
    \caption{Code for defining Fourier continuation.}
    \label{code:wrapper_fno_continuation}
\end{code}

%%%%%%%%%%%%%%%%%%%%%%%%%%%%%%%%%%%%%%%%%
\section{Numerical experiments}\label{section:numerical_experiments}
In this section, we show some numerical examples to demonstrate the effectiveness of our library for the study of neural operators. In particular, we have considered the representative PDE benchmarks provided by \cite{CNO23raonic}, and for more details on these benchmarks, please refer to this reference. These benchmarks are well suited for testing neural operator architectures, providing a large variety of PDEs with intrinsic computational complexity to ensure the effectiveness of NOs. The benchmarks are briefly described below.
\begin{itemize}
    \item \textbf{Poisson equation} benchmark that represents a linear elliptic PDE, given by
          \[ \begin{cases}
                  -\Delta u = f \quad & in \ D         \\
                  u = 0               & on\ \partial D
              \end{cases} \]
          The dataset provides a collection of pairs ${(f^{(i)}, u^{(i)})}_i$ where $u^{(i)}$ are the evaluation of the solution operator $\mathcal{G}: f \to u$ from the source term to the solution.
    \item \textbf{Darcy equation} benchmark that represents a second-order non-linear PDE
          \[ \begin{cases}
                  -\nabla\cdot (a \nabla u) = f  \quad & in \ D\times (0, T) \\
                  u = 0                                & on\ \partial D
              \end{cases} \]
          The dataset provides a collection of pairs ${(a^{(i)}, u^{(i)})}_i$ where $u^{(i)}$ are the evaluation of the solution operator $\mathcal{G}: a \to u$ that maps the diffusion coefficient $a$ into the solution $u$.
    \item \textbf{Navier-Stokes equation} benchmark that represents a non-linear PDE that models the motion of incompressible fluids, given by
          \[ \begin{cases}
                  \frac{\partial u}{\partial t} + (u \cdot \nabla)u + \nabla p = \nu \Delta u \quad & in \ D\times (0, T) \\
                  \nabla \cdot u = 0                                                                & in \ D\times (0, T) \\
                  u(x, 0) = f(x)                                                                    & in\ D               \\
                  \nu = 4 \cdot 10^{-4}
              \end{cases} \]
          The dataset provides a collection of pairs ${(f^{(i)}, u^{(i)})}_i$ where $u^{(i)}$ are the evaluation of the solution operator $\mathcal{G}: f \to u(\cdot, T)$ that maps the initial condition $f$ into the solution at the final time.
    \item \textbf{Wave equation} benchmark that represents a linear hyperbolic PDE, given by
          \[ \begin{cases}
                  \frac{\partial^2u}{\partial t^2} - c^2 \Delta u= 0 \quad & in \ D\times (0, T) \\
                  u(x, 0) = f(x)                                           & in\ D               \\
                  c = 0.1
              \end{cases} \]
          The dataset provides a collection of pairs ${(f^{(i)}, u^{(i)})}_i$ where $u^{(i)}$ are the evaluation of the solution operator $\mathcal{G}: f \to u(\cdot, T)$ that maps the initial condition $f$ into the solution at the final time $T = 5$.
    \item \textbf{Compressible Euler equation} benchmark that represents the following non-linear PDE
          \[ \begin{cases}
                  \frac{\partial u}{\partial t} + \nabla \cdot F(u) = 0 \\
                  u = [\rho,\, \rho v,\, E]^T                           \\
                  F = [\rho v,\, \rho v \otimes v + p\mathbf{I},\, (E + p)v]^T
              \end{cases} \]
          with freestream boundary conditions. The dataset provides a collection of pairs from the shape of the airfoil to the solution.
    \item \textbf{Transport equation} benchmark that represents a non-linear PDE, given by
          \[ \begin{cases}
                  \frac{\partial u}{\partial t}  + v\cdot \nabla u= 0 \quad & in \ D\times (0, T) \\
                  u(x, 0) = f(x)                                            & in\ D               \\
                  v = [0.2, \, 0.2]^T
              \end{cases} \]
          The dataset provides a collection of pairs ${(f^{(i)}, u^{(i)})}_i$ where $u^{(i)}$ are the evaluation of the solution operator $\mathcal{G}: f \to u(\cdot, T)$ that maps the initial condition $f$ into the solution at the final time $T=1$. Two different benchmarks are provided for this PDE, one with smooth initial data and one with discontinuous initial data.
    \item \textbf{Allen-Cahn equation} benchmark that represents a non-linear PDE, given by
          \[ \begin{cases}
                  \frac{\partial u}{\partial t} = \Delta u - \varepsilon^2 u (u^2-1) \quad & in \ D\times (0, T) \\
                  u(x, 0) = f(x)                                                           & in\ D               \\
                  \varepsilon = 220
              \end{cases} \]
          \[
          \]
          The dataset provides a collection of pairs ${(f^{(i)}, u^{(i)})}_i$ where $u^{(i)}$ are the evaluation of the solution operator $\mathcal{G}: f \to u(\cdot, T)$ that maps the initial condition $f$ into the solution at the final time $T=0.0002$.
\end{itemize}

\subsection{Effectiveness of the library}\label{subsection:effectiveness}
For all previous datasets, we train an FNO and a CNO with the choice of hyperparameters reported in the article \cite{CNO23raonic} (and we will call them ``s.o.t.a. hyperparams.'' since they are the best found by the authors) and then we use the hyperparameter optimization workflow (finding our best as possible configuration, called ``our hyperparams'' in this section) and compare the results obtained. The original article shows the results in terms of the relative median $L^1$ loss, instead, we decide to measure the error in the relative mean $L^1$ loss function. Since the hyperparameter optimization process pulls the hyperparameters to obtain larger and heavier architectures, we also run the hyperparameter optimization process fixing the number of trainable parameters to be equal to the s.o.t.a. configuration for a fair comparison, following Section \ref{subsection:same_number_of_params} (we denote these hyperparameter configurations with ``same dof hyperparams'').

For the FNO architecture, we decide to optimize the learning rate, the weight to regularize the loss function ($\lambda\in [10^{-4}, 10^{-2}]$), the factor to reduce the learning rate ($\eta \in [0.75, 0. 99]$), the regularization factor ($w\in [10^{-6}, 10^{-3}]$), the hidden dimension ($d_v\in \{4, 8, 16, 32, 64, 96, 128\}$), the number of hidden layers ($1 \le L \le 6$), the number of modes considered in the Fourier transform ($k_{max}\in \{2, 4, 8, 12, 16, 20, 24, 28, 32\}$), the activation function used ($\sigma \in \{GELU, LReLU, ReLU, tanh\}$ respectively the Gaussian Error Linear Units function \cite{gelu16hen}, LeakyReLU function, Rectified Linear Units function and hyperbolic tangent function), the padding dimension ($0\le pad \le 15$) and the Fourier architecture modification ($arc \in \{Classic, MLP, Residual\}$), see Section \ref{subsection:fno} and \cite{fno24qin}. The resulting errors we get are in Table \ref{table:fno_ray_error} and the hyperparameters found are in Table \ref{table:fno_ray_hyperparams}.
\\
We mention that the best hyperparameters found for the Poisson problem for the FNO architecture (both with the unconstrained optimization and the constrained ``same dofs'' optimization) have only one layer ($L=1$). At first glance, this may seem strange, but it is perfectly interpretable with spectral theory for solving the Poisson problem in the Fourier domain, and is consistent with the proof of the universal approximation theorem for the FNO architecture \cite{uniFNO21kov}.

We performed the same experiment for the emerging CNO model. In this case, we select for optimization the core hyperparameters in the definition of the CNO architecture. In particular, we tune the learning rate ($\lambda \in [10^{-4}, 10^{-2}]$), the learning rate reduction factor ($\eta \in [0.75, 0. 99]$), the regularization factor ($w\in [10^{-6}, 10^{-3}]$), the number of layers ($1\le L\le 5$), the number of layers for the final residual block ($1 \le N \le 6$), the number of layers for the remaining residual blocks ($1 \le M\le 8$), the channel multiplier ($chan\_mul\in \{8, 16, 24, 32, 40, 48\}$) and the kernel size ($k \in \{3, 5, 7\}$). The resulting errors are listed in Table \ref{table:cno_ray_error} and the hyperparameters are listed in Table \ref{table:cno_ray_hyperparams}.

For all the tests performed, we use the AdamW optimizer, reducing the learning rate by a factor of $\eta$ every $10$ epoch for the FNO architecture and every epoch for the CNO architecture.
We try for $200$ hyperparameters configuration, the time needed for each test strongly depends on the dimension of the datasets and the configuration of the search space, our tests take up to $2$ days each on a single $4090$ NVIDIA GPU.

\begin{table}[!ht] %booktabs
    \centering
    \begin{tabular}{cccc}\toprule
        problem                     & s.o.t.a. hyperparams. & same dofs hyperparams.          & our hyperparams.                   \\
        \cmidrule{1-4}
        Poisson eq.                 & $ (5.61\pm 0.094) \%$ & $ (1.46 \pm 0.071) \% $         & $ \mathbf{(1.04 \pm 0.090)\%} $    \\
        Darcy eq.                   & $(1.05 \pm 0.008) \%$ & $ \mathbf{(0.81\pm 0.015)} \% $ & $ (1.00 \pm 0.006) \% $            \\
        Navier-Stokes eq.           & $ (3.83\pm 0.096) \%$ & $ (3.26 \pm 0.083) \% $         & $ \mathbf{(3.26 \pm 0.083)} \% $   \\
        Wave eq.                    & $ (1.27\pm 0.048) \%$ & $ \mathbf{(0.95\pm 0.094)} \% $ & $ (1.08 \pm 0.038) \% $            \\
        Euler eq.                   & $ (0.48\pm 0.017) \%$ & $ (0.42 \pm 0.002) \% $         & $ \mathbf{(0.42 \pm 0.002)} \% $   \\
        Smooth transport eq.        & $ (0.40\pm 0.026) \%$ & $ (0.17\pm 0.006) \% $          & $ \mathbf{(0.036 \pm 0.0004)} \% $ \\
        Discontinuous transport eq. & $ (1.30\pm 0.080) \%$ & $ (1.09\pm 0.071) \% $          & $ \mathbf{(1.01 \pm 0.018)} \% $   \\
        Allen-Cahn eq.              & $ (0.48\pm 0.052)\%$  & $ \mathbf{(0.22\pm 0.006)} \% $ & $ (0.25 \pm 0.028)\%$              \\
        \bottomrule
    \end{tabular}
    \caption{The tests are performed for all the benchmarks previously described with the FNO architecture. In the table we report the resulting mean $L^1$ errors obtained with the s.o.t.a. hyperparameters (second column), the optimized hyperparameter configuration (last column) and the hyperparameters with the constraint to keep the number of trainable parameters fixed and equal to the s.o.t.a. configuration (third column). For each test, we report the mean and the standard deviation after three trials.}
    \label{table:fno_ray_error}
\end{table}
\begin{table}[!ht] %booktabs
    \centering
    \begin{tabular}{ccccccccccc}\toprule
         & problem                     & $\lambda$ & $\eta$  & $w$                 & $d_v$ & $L$ & $k_{max}$ & $\sigma$ & pad  & arc      \\
        \cmidrule{1-11}
        \multirow{8}{*}{\rotatebox[origin=c]{90}{\footnotesize free hyperparams.}}
         & Poisson eq.                 & $0.0005$  & $0.83$  & $0.1\cdot 10^{-4}$  & $128$ & $1$ & $32$      & tanh     & $0$  & MLP      \\
         & Darcy eq.                   & $0.0003$  & $0.94$  & $0.8\cdot 10^{-4}$  & $32$  & $4$ & $32$      & LReLU    & $0$  & Classic  \\
         & Navier-Stokes eq.           & $0.00308$ & $0.93$  & $6.75\cdot 10^{-4}$ & $64$  & $5$ & $20$      & LReLU    & $0$  & MLP      \\
         & Wave eq.                    & $0.00173$ & $0.94$  & $0.03\cdot 10^{-4}$ & $32$  & $4$ & $28$      & GELU     & $1$  & Classic  \\
         & Euler eq.                   & $0.008$   & $0.88$  & $3.45\cdot 10^{-4}$ & $32$  & $3$ & $8$       & GELU     & $9$  & MLP      \\
         & Smooth transport eq.        & $0.00298$ & $0.9$   & $7.51\cdot 10^{-4}$ & $128$ & $5$ & $28$      & LReLU    & $5$  & Classic  \\
         & Disc. transport eq.         & $0.00681$ & $0.79$  & $1.77\cdot 10^{-4}$ & $64$  & $4$ & $32$      & tanh     & $0$  & Residual \\
         & Allen-Cahn eq.              & $0.00275$ & $0.755$ & $2.38\cdot 10^{-4}$ & $8$   & $2$ & $28$      & ReLU     & $3$  & Classic  \\
        \cmidrule{1-11}
        \multirow{6}{*}{\rotatebox[origin=c]{90}{\footnotesize same dofs}}
         & Poisson           & $0.00721$ & $0.96$  & $0.12\cdot 10^{-4}$ & $8$   & $1$ & $16$      & LReLU    & $2$  & Classic  \\
         & Darcy             & $0.00104$ & $0.97$  & $2.5\cdot 10^{-4}$  & $32$  & $3$ & $24$      & GELU     & $14$ & MLP      \\
         & Wave              & $0.00223$ & $0.94$  & $7.11\cdot 10^{-4}$ & $16$  & $3$ & $20$      & GELU     & $6$  & Classic  \\
         & Smooth transport  & $0.0081$  & $0.86$  & $9.07\cdot 10^{-4}$ & $16$  & $5$ & $20$      & LReLU    & $5$  & MLP      \\
         & Disc. transport   & $0.00852$ & $0.88$  & $5.86\cdot 10^{-4}$ & $32$  & $3$ & $16$      & tanh     & $5$  & Residual \\
         & Allen-Cahn        & $0.00702$ & $0.96$  & $6.5\cdot 10^{-4}$  & $16$  & $1$ & $20$      & LReLU    & $7$  & Classic  \\
        \bottomrule
    \end{tabular}
    \caption{Optimized hyperparameters for FNO architecture across all the benchmarks. Where the hyperparameter for the ``same dofs'' are not specified means that they are equals to the ``free hyperparameters''.}
    \label{table:fno_ray_hyperparams}
\end{table}
\begin{table}[!ht] %booktabs
    \centering
    \begin{tabular}{cccc}\toprule
        problem                     & s.o.t.a. hyperparams.  & same dofs hyperparams.       & our hyperparams.                 \\
        \cmidrule{1-4}
        Poisson eq.                 & $ (0.45 \pm 0.017) \%$ & $\mathbf{(0.30\pm 0.035)}\%$ & $ (0.31 \pm 0.036)\% $           \\
        Darcy eq.                   & $ (0.67 \pm 0.009) \%$ & $ (0.56 \pm 0.035) \% $      & $ \mathbf{(0.56 \pm 0.035)} \% $ \\
        Navier-Stokes eq.           & $(3.67 \pm 0.170) \%$  & $(3.57\pm 0.106)\%$          & $ \mathbf{(3.45 \pm 0.154)} \% $ \\
        Wave eq.                    & $ (1.54 \pm 0.015) \%$ & $(1.38\pm 0.099)\%$          & $ \mathbf{(0.96 \pm 0.060)} \% $ \\
        Euler eq.                   & $ (0.43\pm 0.007) \%$  & $ (0.39 \pm 0.001) \% $      & $ \mathbf{(0.39 \pm 0.001)} \% $ \\
        Smooth transport eq.        & $ (0.84\pm 0.041) \%$  & $(0.47\pm 0.049)\%$          & $ \mathbf{(0.42 \pm 0.052)} \% $ \\
        Discontinuous transport eq. & $ (1.41 \pm 0.006) \%$ & $(1.21\pm 0.049)\%$          & $ \mathbf{(1.21 \pm 0.025)} \% $ \\
        Allen-Cahn eq.              & $ (2.06\pm 0.254)\%$   & $(1.43\pm 0.102)\%$          & $ \mathbf{(1.41 \pm 0.033)}\%$   \\
        \bottomrule
    \end{tabular}
    \caption{The tests are performed for all the benchmarks previously described with the CNO architecture. In the table we report the resulting mean $L^1$ errors obtained with the s.o.t.a. hyperparameters (second column), the optimized hyperparameter configuration (last column) and the hyperparameters with the constraint to keep the number of trainable parameters fixed and equal to the s.o.t.a. configuration (third column). For each test, we report the mean and the standard deviation after three trials.}
    \label{table:cno_ray_error}
\end{table}
\begin{table}[!ht] %booktabs
    \centering
    \begin{tabular}{cccccccccc}\toprule
         & problem              & $\lambda$ & $\eta$ & $w$                 & $L$ & $N$ & $M$ & chan\_mul & ker\_size \\
        \cmidrule{1-10}
        \multirow{8}{*}{\rotatebox[origin=c]{90}{\footnotesize free hyperparams.}}
         & Poisson eq.          & $0.00595$ & $0.97$ & $8.18\cdot 10^{-4}$ & $3$ & $1$ & $2$ & $16$      & $5$       \\
         & Darcy eq.            & $0.0056$  & $0.99$ & $8.8\cdot 10^{-4}$  & $3$ & $3$ & $3$ & $40$      & $3$       \\
         & Navier-Stokes eq.    & $0.00218$ & $0.99$ & $5.17\cdot 10^{-4}$ & $4$ & $3$ & $7$ & $56$      & $3$       \\
         & Wave eq.             & $0.00282$ & $0.99$ & $0.51\cdot 10^{-4}$ & $4$ & $3$ & $7$ & $32$      & $5$       \\
         & Euler eq.            & $0.00377$ & $0.99$ & $4.56\cdot 10^{-4}$ & $3$ & $5$ & $2$ & $48$      & $3$       \\
         & Smooth transport eq. & $0.00211$ & $0.99$ & $8.83\cdot 10^{-4}$ & $2$ & $5$ & $4$ & $32$      & $7$       \\
         & Disc. transport eq.  & $0.00622$ & $0.99$ & $3.0\cdot 10^{-4}$  & $3$ & $3$ & $1$ & $56$      & $3$       \\
         & Allen-Cahn eq.       & $0.0016$  & $0.99$ & $4.25\cdot 10^{-4}$ & $3$ & $3$ & $4$ & $48$      & $3$       \\
        \cmidrule{1-10}
        \multirow{6}{*}{\rotatebox[origin=c]{90}{\footnotesize same dofs}}
         & Poisson eq.          & $0.00147$ & $0.98$ & $4.81\cdot 10^{-4}$ & $2$ & $2$ & $6$ & $40$      & $5$       \\
         & Navier-Stokes eq.    & $0.00129$ & $0.99$ & $2.3\cdot 10^{-4}$  & $3$ & $3$ & $6$ & $23$      & $5$       \\
         & Wave eq.             & $0.00369$ & $0.99$ & $3.6\cdot 10^{-4}$  & $4$ & $4$ & $3$ & $27$      & $3$       \\
         & Smooth transport eq. & $0.00138$ & $0.99$ & $4.6\cdot 10^{-4}$  & $3$ & $3$ & $5$ & $33$      & $3$       \\
         & Disc. transport eq.  & $0.0093$  & $0.99$ & $3.8\cdot 10^{-5}$  & $4$ & $1$ & $4$ & $17$      & $3$       \\
         & Allen-Cahn eq.       & $0.00084$ & $0.99$ & $2.7\cdot 10^{-4}$  & $3$ & $2$ & $4$ & $32$      & $3$       \\
        \bottomrule
    \end{tabular}
    \caption{Optimized hyperparameters for CNO architecture across all the benchmarks. Where the hyperparameter for the ``same dofs'' are not specified means that they are equals to the ``free hyperparameters''.}
    \label{table:cno_ray_hyperparams}
\end{table}

\subsection{Smooth transport equation example}
\subsubsection*{Hyperparameter configurations comparison}
\begin{figure}[!ht]
    \centering
    \begin{subfigure}{\textwidth}
        \centering
        \includegraphics[width=\textwidth]{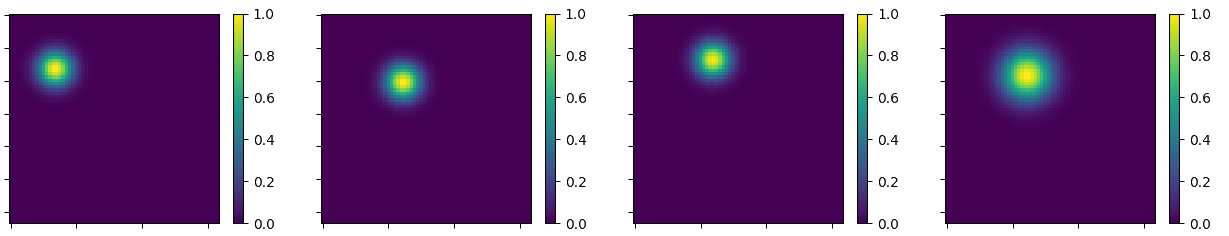}
        \caption{Input smooth functions.}
        \label{fig:input}
    \end{subfigure}
    \vfill
    \begin{subfigure}{\textwidth}
        \centering
        \includegraphics[width=\textwidth]{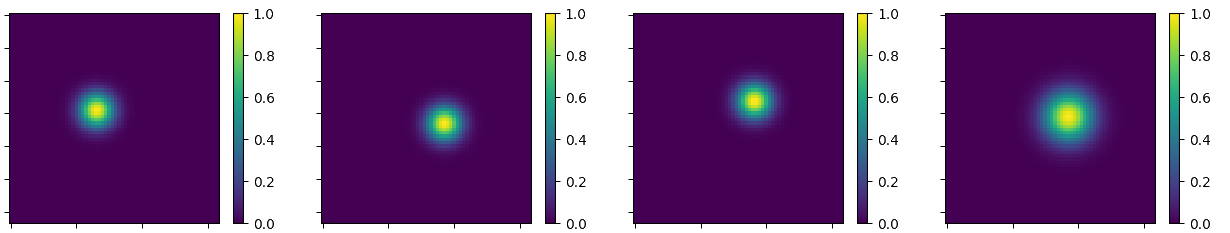}
        \caption{High fidelity solutions.}
        \label{fig:output}
    \end{subfigure}
    \vfill
    \begin{subfigure}{\textwidth}
        \centering
        \includegraphics[width=\textwidth]{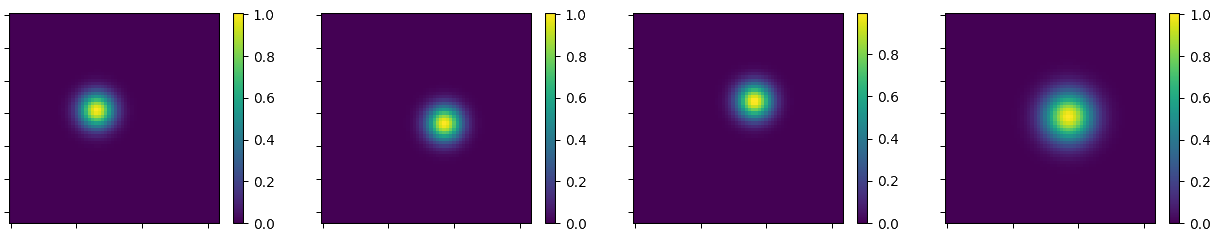}
        \caption{FNO approximations with the s.o.t.a. hyperparameters configuration.}
        \label{fig:appro_default}
    \end{subfigure}
    \vfill
    \begin{subfigure}{\textwidth}
        \centering
        \includegraphics[width=\textwidth]{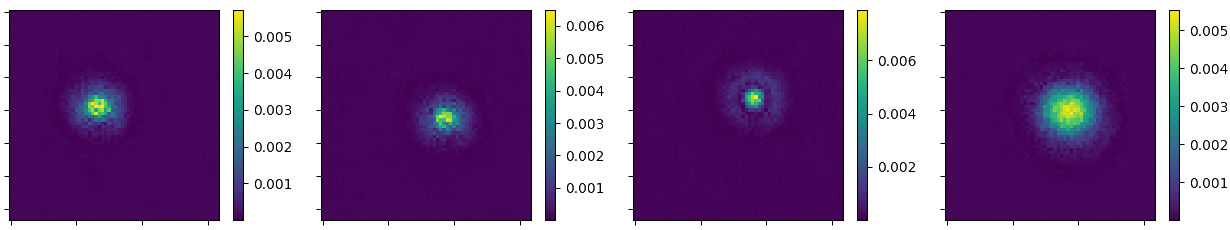}
        \caption{Approximation error with the s.o.t.a. hyperparameters configuration.}
        \label{fig:error_default}
    \end{subfigure}
    \vfill
    \begin{subfigure}{\textwidth}
        \centering
        \includegraphics[width=\textwidth]{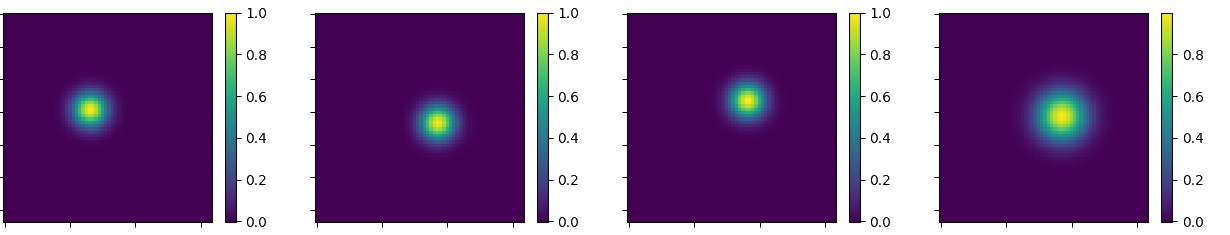}
        \caption{FNO approximations with the best hyperparameters configuration found.}
        \label{fig:appro_best}
    \end{subfigure}
    \vfill
    \begin{subfigure}{\textwidth}
        \centering
        \includegraphics[width=\textwidth]{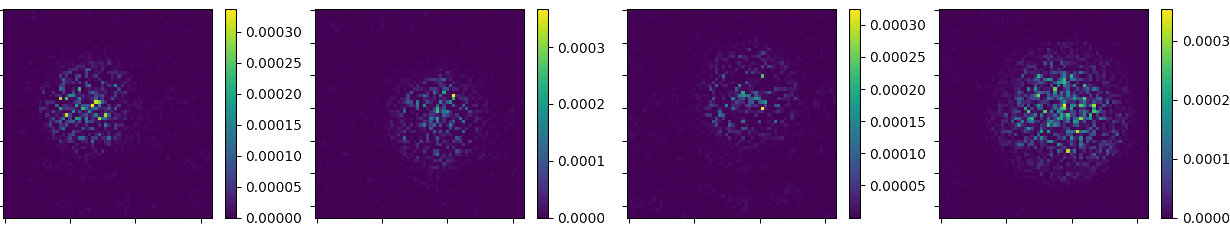}
        \caption{Approximation error with the best hyperparameters configuration found.}
        \label{fig:error_best}
    \end{subfigure}
    \caption{We represent four of the input functions (Fig. \ref{fig:input}) where the default parameters configuration gets the higher error, with the corresponding true output (Fig.\ref{fig:output}), the FNO's approximation (Fig. \ref{fig:appro_default}) and the absolute value of the error (Fig. \ref{fig:error_default}) with the previous s.o.t.a. hyperparameters. Then, for the same input, we compute and plot the FNO's approximation obtained with our optimized hyperparameter configuration that we found (Fig. \ref{fig:appro_best}) with the corresponding absolute error (Fig. \ref{fig:error_best}).}
    \label{fig:FNO_default_best}
\end{figure}

The most interesting case is the transport equation with smooth initial data, where we can see from Table \ref{table:fno_ray_error} that careful tuning of the hyperparameters of the FNO's architecture can lead to an order of magnitude improvement in the loss function, eliminating errors due to manual optimization using a heuristic hyperparameter optimization process based on grid search or random search algorithms. For better understanding, we examine four input functions where the default parameter configuration results in the highest error. For each input function, we show the true output, the approximation of the FNO under the default configurations, and the absolute error between the true output and the approximation of the FNO. For the same input functions, we also compute and plot the FNO's approximation using the optimized hyperparameter configuration obtained through our optimization process, alongside the corresponding absolute error in Figure \ref{fig:FNO_default_best}. From the plots, we can see that the optimized hyperparameter configuration outperforms the default configuration in all four examples, with more spread and smaller absolute values of the error. Here we only plot four random examples in this interesting case, but you can find all the trained models, evaluated and plotted along all the test sets, at the following website \href{https://hypernos.streamlit.app/}{https://hypernos.streamlit.app/}.

\begin{figure}[!ht]
    \centering
    \begin{subfigure}[t]{0.49\textwidth}
        \centering
        \includegraphics[width=\textwidth, height=0.181\textheight]{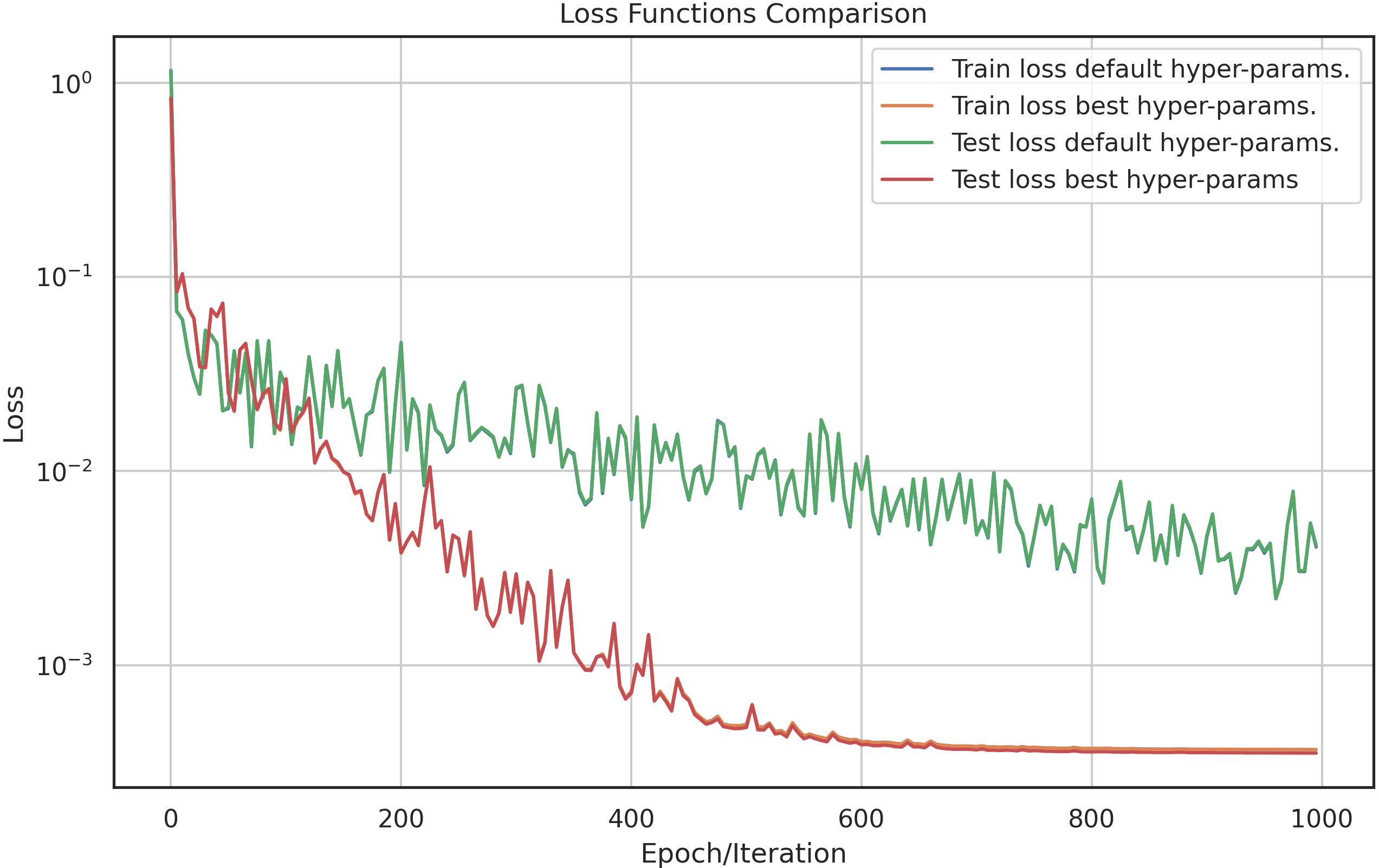}
        \caption{Plot of loss functions.}
        \label{fig:loss_fun}
    \end{subfigure}%
    \hfill%
    \begin{subfigure}[t]{0.49\textwidth}
        \centering
        \includegraphics[width=\textwidth]{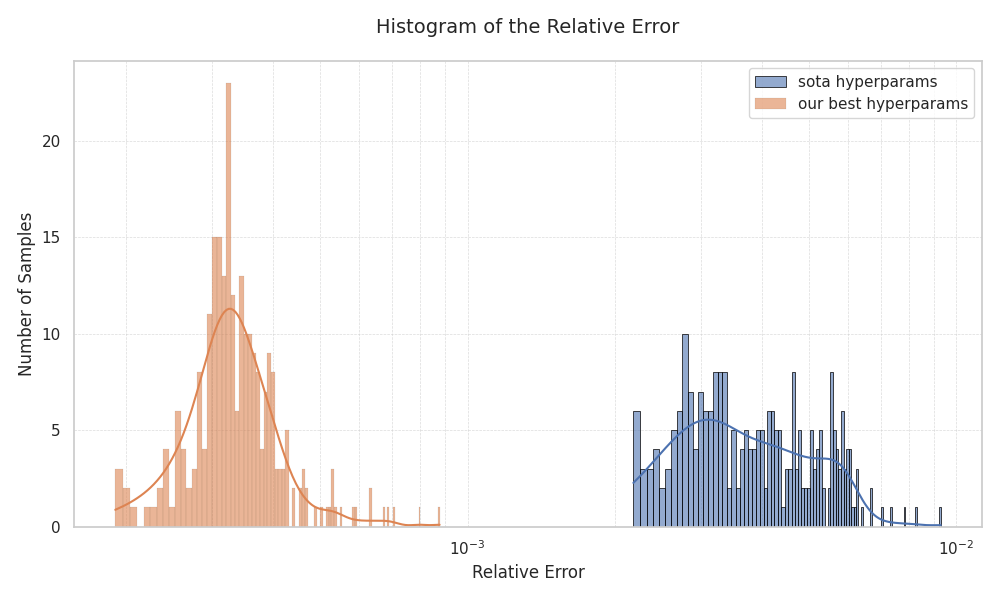}
        \caption{Histogram of distribution of the test errors.}
        \label{fig:distribution}
    \end{subfigure}%
    \hfill%
    \begin{subfigure}[t]{0.49\textwidth}
        \centering
        \includegraphics[width=\textwidth, height=0.181\textheight]{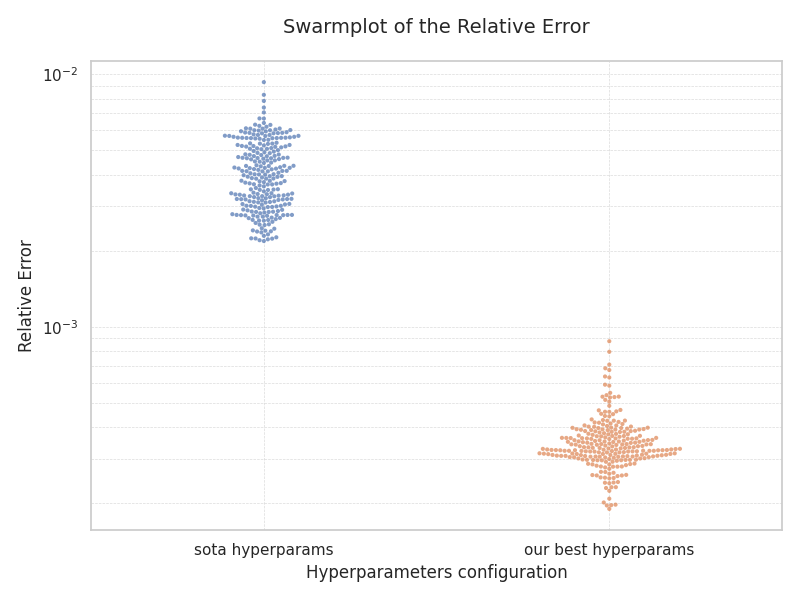}
        \caption{Swarm-plot of distribution of the test errors.}
        \label{fig:swarm}
    \end{subfigure}
    \hfill%
    \caption{We plot the loss functions (relative $L^1$ error) with respect to the train set and the test set for both the default and the best hyperparameter configurations (Fig. \ref{fig:loss_fun}) and display the distribution of the relative $L^1$ error with histograms (Fig. \ref{fig:distribution}) and warm-plot (Fig. \ref{fig:swarm}). In Figure \ref{fig:loss_fun} the train and the test are almost overlapping, so we can see only one line.}
    \label{fig:FNO_comparison}
\end{figure}

Furthermore, we plot the loss functions for both the train set and the test set, comparing the results for the default hyperparameter configuration and the optimized (best) configuration. We also provide a plot showing the distribution of the error over the test set, see Figure \ref{fig:FNO_comparison}. From Figure \ref{fig:loss_fun} we can see that the training loss function with the optimized hyperparameters decreases faster and with less oscillation at the end of the training process. We can also see from Figures \ref{fig:distribution}, \ref{fig:swarm} that the error of the model with optimized hyperparameters is consistently one order of magnitude lower than the error of the model with the default configuration.

This comprehensive comparison highlights the improved accuracy and performance achieved by the optimized hyperparameters consistently across all quantities considered.

\begin{figure}[!ht]
    \centering
    \includegraphics[width=\textwidth]{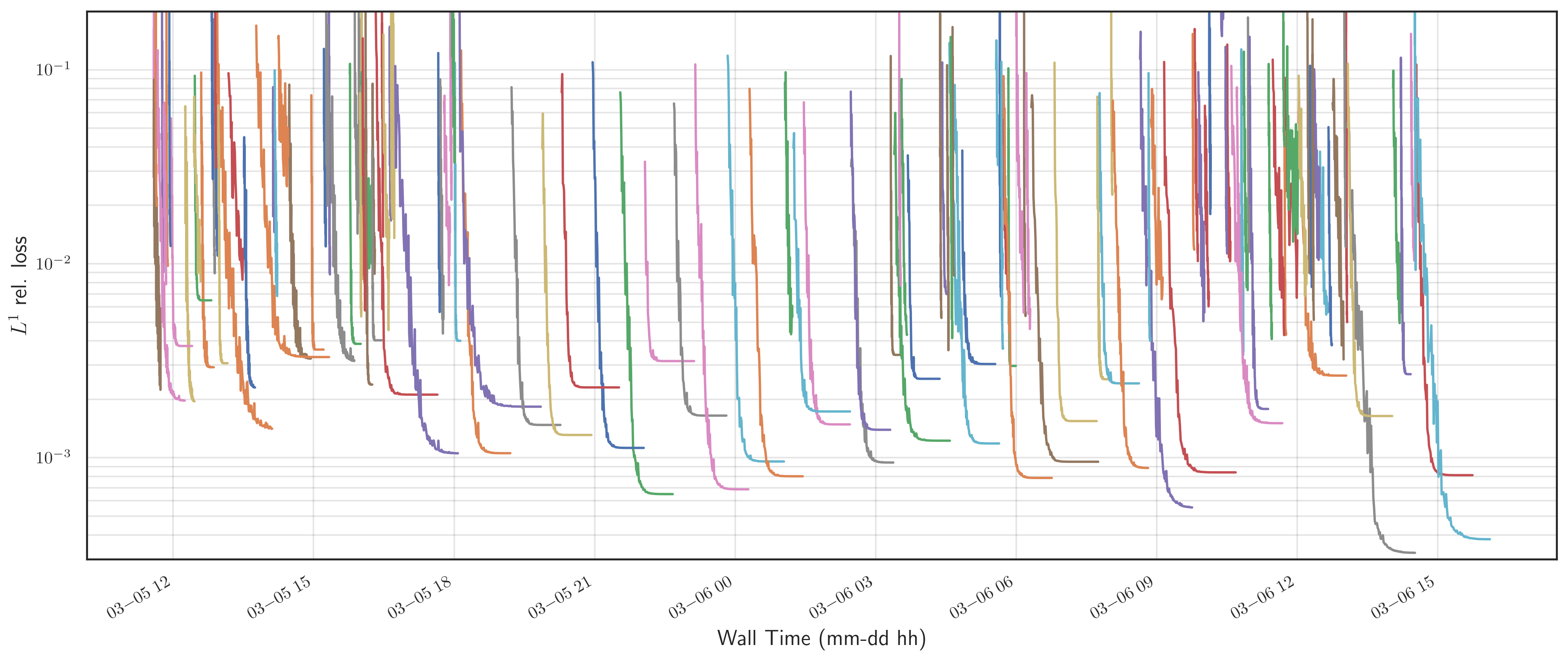}
    \caption{In this plot we can see the relative $L^1$ error of each hyperparameter configuration with respect to the passed time from the beginning of the simulation.}
    \label{fig:wall_time_loss}
\end{figure}

We also plot the \textit{wall time plot}, see Figure \ref{fig:wall_time_loss}. This plot shows the loss functions for all hyperparameter configurations tested during the optimization process for the FNO model approximating the operator associated with the transport equation with smooth data. The plot has on the y-axis the relative $L^1$ error and on the x-axis the time (the whole simulation takes about $27$ hours) and tries with $150$ different hyperparameter configurations. In other words, this plot represents the best relative error in the loss function that we get as the simulation time increases. The first loss plotted is the loss relative to the default hyperparameter configuration, i.e. the first hyperparameter configuration tested, as mentioned in Section \ref{subsection:default_hyperparams}. For this test, we use a single GPU but run two trials in parallel, which explains why there are two overlapping losses at the same time. The \textit{width} of a single loss is an indicator of the weight and size of the model. We can also see that the best configuration is obtained almost at the end of the optimization process, but near-optimal results are already obtained in the first half of the simulation.

\subsubsection*{Same complexity configurations}
\begin{figure}
\centering
\includegraphics[width=0.5\textwidth]{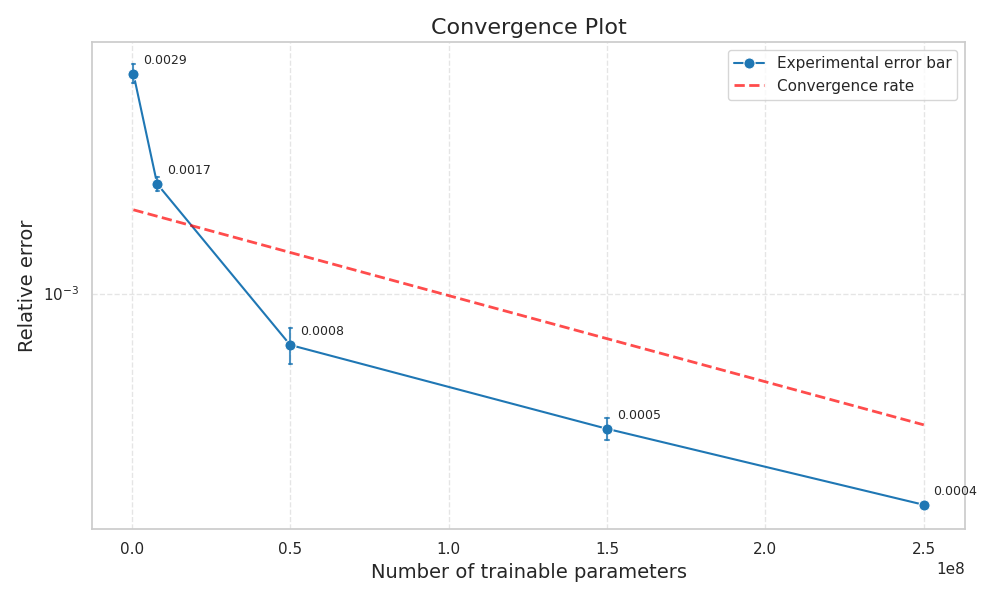}
\caption{Here we plot the convergence of the loss function (relative $L^1$ error) evaluated on the test set for the FNO architecture approximating the transport equation with continuous initial data with respect to the number of trainable parameters of the architecture. Near the dot we report the standard deviation of the error.}
\label{figure:conv_plot}
\end{figure}

From the previous comparison, one can see from Table \ref{table:fno_ray_hyperparams} that the optimized model have a big amount of trainable parameters, in particular, our optimized results have $250^\cdot 000 ^ \cdot 000$ parameters instead the s.o.t.a. model has $8^\cdot 000 ^ \cdot 000$ parameters. So we can study how much we can improve the solution with the same number of trainable parameters. This is done with the routine explained in \ref{subsection:same_number_of_params}, and the obtained results are reported in Table \ref{table:fno_ray_hyperparams}. We can see that in this example we can improve the performance by about a factor of two even with the same number of trainable parameters, but to unlock all the potential we have to explore even bigger and more complicated models. So we can study how much we can improve the solution with the same number of trainable parameters, and the obtained results are reported in Table \ref{table:fno_ray_error_samedof}; where we report the previous state of the art results (called ``s.o.t.a.''), the best that we can do with the same number of trainable parameters (called ``best samedofs''), the best configuration with unconstrained optimization (called ``best'') and the best that we can do with a fixed maximum value of trainable parameters respectively $500^\cdot 000$, $50^\cdot 000 ^ \cdot 000$ and $150^\cdot 000 ^ \cdot 000$ (called ``best $500k$'', ``best $50M$'', ``best $150 M$'' respectively). In Table \ref{table:fno_ray_hyperparams_dof} we report the hyperparameters found for the new configurations. Finally in Figure \ref{figure:conv_plot} we plot the convergence of the loss function (relative $L^1$ error) evaluated on the test set for the optimized FNO architecture with respect to the number of trainable parameters of the architecture, as we can see the error is decreasing when the number of trainable parameters increases.
\begin{table}[!ht] %booktabs
    \centering
    \footnotesize
    \begin{tabular}{cccccc}\toprule
        s.o.t.a  & best $500k$ & best samedofs ($8$M) & best $50M$ & best $150M$ & best ($250M$)\\
        \cmidrule{1-6}
        $ (0.40\pm 0.026) \%$ & $(0.29 \pm 0.013) \%$ & $ (0.17 \pm 0.006) \% $ & $ (0.078 \pm 0.0067) \% $ & $ (0.052 \pm 0.0027) \% $ & $ \mathbf{(0.036 \pm 0.0004) \%} $  \\
        \bottomrule
    \end{tabular}
    \caption{With the notation previously introduced for the different FNO hyperparameters optimization, in this table we report the mean and standard deviation for the testing error calculated with three different trials.}
    \label{table:fno_ray_error_samedof}
\end{table}
\begin{table}[!ht] %booktabs
    \centering
    \begin{tabular}{cccccccccc}\toprule
        problem   & $\lambda$ & $\eta$ & $w$                 & $d_v$ & $L$ & $k_{max}$ & $\sigma$ & pad & arc     \\
        \cmidrule{1-10}
        best $500k$ & $0.00455$ & $0.93$  & $7.0\cdot 10^{-4}$ & $16$ & $4$ & $11$      & LReLU    & $7$ & Classic \\
        best $50M$ & $0.0078$ & $0.89$ & $0.67\cdot 10^{-4}$ & $64$  & $5$ & $24$      & LReLU    & $12$ & Classic\\
        best $150M$ & $0.00874$ & $0.89$ & $1.12\cdot 10^{-4}$ & $48$  & $5$ & $32$      & LReLU    & $15$ & Classic\\
        \bottomrule
    \end{tabular}
    \caption{We write the optimized hyperparameters for FNO architecture on the transport equation with continuous initial data. In particular, we report the hyperparameters found for the new configurations, the others are reported in Table \ref{table:fno_ray_hyperparams}.}
    \label{table:fno_ray_hyperparams_dof}
\end{table}

%%%%%%%%%%%%%%%%%%%%%%%%%%%%%%%%%%%%%%%%%
\section{Conclusions}\label{section:conclusions}
The HyperNOs library is an important and useful tool for the study of neural operators, providing a powerful and scalable framework for automated hyperparameter optimization. By combining parallel computing with state-of-the-art optimization algorithms, HyperNOs enables researchers and practitioners to efficiently tune the hyperparameters of neural operators. This is useful for unlocking the full potential of neural operators to solve complex real-world problems, for exhaustively exploring the properties of neural operators, and for fair and comprehensive comparison of different architectures and models. The library's user-friendly interface, high flexibility in both data set and model definition, high-level functionality, and integration with popular neural operator frameworks make it a helpful tool for advancing research and applications in this rapidly evolving field. By automating this critical aspect of model development, researchers and practitioners can focus on higher-level tasks such as designing novel architectures, interpreting results, and applying neural operators to solve new and challenging real-world problems.

In future work, we aim to expand the range of implemented architectures available in the library, thereby increasing its versatility and applicability to a wider range of problems. This expansion will include the integration of state-of-the-art neural operator architectures, such as deep operator network, graph neural operators and attention-based models, which have shown significant success in various domains. To further empower users of the library, we intend to introduce new functionalities that streamline the development and deployment of neural operator models. Through these enhancements, we aim to position the library as a comprehensive and user-friendly platform that not only meets the current needs of the neural operator and deep learning community, but also anticipates and adapts to future advances in the field. By fostering collaboration and encouraging contributions from the open-source community, we aim to ensure that the library remains at the forefront of innovation, continuously evolving to meet the challenges and opportunities of modern machine learning.

\subsection*{Code availability}
The \textbf{library} is public on GitHub and can be downloaded at \url{https://github.com/MaxGhi8/HyperNOs} following the related instructions. In addition to the library, we provide a \textbf{web application} that allows users to interact with the library results and visualize it for all the described models across the test sets of all the benchmarks. The web application is available at \url{https://hypernos.streamlit.app/} and the source code is available at \url{https://github.com/MaxGhi8/HyperNOs_website}. Furthermore, all the \textbf{trained models} are made public on Zenodo at \cite{Max25models}.

\subsection*{Acknowledgements}
The author would like to thank Siddhartha Mishra for useful discussions and suggestions, and Giorgio Abelli for valuable support in refactoring some parts of the library.

\clearpage


\clearpage
\begin{thebibliography}{99}

    \bibitem{moller24iganets}
    Backmeyer, Merle, Stefan Kurz, Matthias Möller, and Sebastian Schöps. ``Solving Electromagnetic Scattering Problems by Isogeometric Analysis with Deep Operator Learning.'' In 2024 Kleinheubach Conference, pp. 1-4. IEEE, 2024.

    \bibitem{reno24bar}
    Bartolucci, Francesca, Emmanuel de Bézenac, Bogdan Raonic, Roberto Molinaro, Siddhartha Mishra, and Rima Alaifari. ``Representation equivalent neural operators: a framework for alias-free operator learning.'' Advances in Neural Information Processing Systems 36 (2024).

    \bibitem{hyperopt11bergstra}
    Bergstra, James, Rémi Bardenet, Yoshua Bengio, and Balázs Kégl. ``Algorithms for hyper-parameter optimization.'' Advances in neural information processing systems 24 (2011).

    \bibitem{spectral06canuto}
    Canuto, Claudio, M. Youssuff Hussaini, Alfio Quarteroni, and Thomas A. Zang. Spectral methods. Vol. 285. Berlin: springer, 2006.

    \bibitem{transformer21cao}
    Cao, Shuhao. ``Choose a transformer: Fourier or galerkin.'' Advances in neural information processing systems 34 (2021): 24924-24940.

    \bibitem{physloss24deryck}
    De Ryck, Tim, and Siddhartha Mishra. ``Numerical analysis of physics-informed neural networks and related models in physics-informed machine learning.'' arXiv preprint arXiv:2402.10926 (2024).

    \bibitem{SNO23fanaskov}
    Fanaskov, Vladimir Sergeevich, and Ivan V. Oseledets. ``Spectral neural operators.'' In Doklady Mathematics, vol. 108, no. Suppl 2, pp. S226-S232. Moscow: Pleiades Publishing, 2023.

    \bibitem{Max25models}
    Ghiotto, Massimiliano. “CNO and FNO Trained Model Across the Representative PDE Benchmarks”. Zenodo, March 20, 2025. \url{https://doi.org/10.5281/zenodo.15055547.}

    \bibitem{Max25FNOvideo}
    Ghiotto, Massimiliano. Visual Representation of the Fourier Neural Operator Architecture in 1D and 2D. Zenodo, 2025. \url{https://doi.org/10.5281/zenodo.14884675.}

    \bibitem{gelu16hen}
    Hendrycks, Dan, and Kevin Gimpel. ``Gaussian error linear units (gelus).'' arXiv preprint arXiv:1606.08415 (2016).

    \bibitem{poseidon24her}
    Herde, Maximilian, Bogdan Raonić, Tobias Rohner, Roger Käppeli, Roberto Molinaro, Emmanuel de Bézenac, and Siddhartha Mishra. Poseidon: Efficient Foundation Models for PDEs. arXiv preprint arXiv:2405.19101 (2024).

    \bibitem{iga05hughes}
    Hughes, Thomas JR, John A. Cottrell, and Yuri Bazilevs. ``Isogeometric analysis: CAD, finite elements, NURBS, exact geometry and mesh refinement.'' Computer methods in applied mechanics and engineering 194, no. 39-41 (2005): 4135-4195.

    \bibitem{uniFNO21kov}
    Kovachki, Nikola, Samuel Lanthaler, and Siddhartha Mishra. ``On universal approximation and error bounds for Fourier neural operators.'' The Journal of Machine Learning Research 22, no. 1 (2021): 13237-13312.

    \bibitem{neuraloperator21kov}
    Kovachki, Nikola, Zongyi Li, Burigede Liu, Kamyar Azizzadenesheli, Kaushik Bhattacharya, Andrew Stuart, and Anima Anandkumar. Neural operator: Learning maps between function spaces with applications to pdes. Journal of Machine Learning Research 24, no. 89 (2023): 1-97.

    \bibitem{don22lan}
    Lanthaler, Samuel, Siddhartha Mishra, and George E. Karniadakis. ``Error estimates for deeponets: A deep learning framework in infinite dimensions.'' Transactions of Mathematics and Its Applications 6, no. 1 (2022): tnac001.

    \bibitem{nno23lan}
    Lanthaler, Samuel, Zongyi Li, and Andrew M. Stuart. ``The nonlocal neural operator: Universal approximation.'' arXiv preprint arXiv:2304.13221 (2023).

    \bibitem{seismicFNO23li}
    Li, Bian, Hanchen Wang, Shihang Feng, Xiu Yang, and Youzuo Lin. `'`Solving seismic wave equations on variable velocity models with Fourier neural operator.'' IEEE Transactions on Geoscience and Remote Sensing 61 (2023): 1-18.

    \bibitem{pino24li}
    Li, Zongyi, Hongkai Zheng, Nikola Kovachki, David Jin, Haoxuan Chen, Burigede Liu, Kamyar Azizzadenesheli, and Anima Anandkumar. ``Physics-informed neural operator for learning partial differential equations.'' ACM/JMS Journal of Data Science 1, no. 3 (2024): 1-27.

    \bibitem{asha20li}
    Li, Liam, Kevin Jamieson, Afshin Rostamizadeh, Ekaterina Gonina, Jonathan Ben-Tzur, Moritz Hardt, Benjamin Recht, and Ameet Talwalkar. ``A system for massively parallel hyperparameter tuning.'' Proceedings of Machine Learning and Systems 2 (2020): 230-246.

    \bibitem{MGNO20li}
    Li, Zongyi, Nikola Kovachki, Kamyar Azizzadenesheli, Burigede Liu, Andrew Stuart, Kaushik Bhattacharya, and Anima Anandkumar. ``Multipole graph neural operator for parametric partial differential equations.'' Advances in Neural Information Processing Systems 33 (2020): 6755-6766.

    \bibitem{GNO20li}
    Li, Zongyi, Nikola Kovachki, Kamyar Azizzadenesheli, Burigede Liu, Kaushik Bhattacharya, Andrew Stuart, and Anima Anandkumar. ``Neural operator: Graph kernel network for partial differential equations.'' arXiv preprint arXiv:2003.03485 (2020).

    \bibitem{FNO20li}
    Li, Zongyi, Nikola Kovachki, Kamyar Azizzadenesheli, Burigede Liu, Kaushik Bhattacharya, Andrew Stuart, and Anima Anandkumar, Fourier neural operator for parametric partial differential equations, arXiv preprint arXiv:2010.08895 (2020).

    \bibitem{GIFNO24li}
    Li, Zongyi, Nikola Kovachki, Chris Choy, Boyi Li, Jean Kossaifi, Shourya Otta, Mohammad Amin Nabian et al. ``Geometry-informed neural operator for large-scale 3d pdes.'' Advances in Neural Information Processing Systems 36 (2024).

    \bibitem{OPNO24liu}
    Liu, Ziyuan, Haifeng Wang, Hong Zhang, Kaijun Bao, Xu Qian, and Songhe Song. ``Render unto Numerics: Orthogonal Polynomial Neural Operator for PDEs with Nonperiodic Boundary Conditions.'' SIAM Journal on Scientific Computing 46, no. 4 (2024): C323-C348.

    \bibitem{DON21lu}
    Lu, Lu, Pengzhan Jin, Guofei Pang, Zhongqiang Zhang, and George Em Karniadakis. Learning nonlinear operators via DeepONet based on the universal approximation theorem of operators. Nature Machine Intelligence 3, no. 3 (2021): 218-229.

    \bibitem{don23mar}
    Marcati, C., Schwab, C. (2023). Exponential convergence of deep operator networks for elliptic partial differential equations. SIAM Journal on Numerical Analysis, 61(3), 1513-1545.

    \bibitem{ray18moritz}
    Moritz, Philipp, Robert Nishihara, Stephanie Wang, Alexey Tumanov, Richard Liaw, Eric Liang, Melih Elibol et al. ``Ray: A distributed framework for emerging {AI} applications.'' In 13th USENIX symposium on operating systems design and implementation (OSDI 18), pp. 561-577. 2018.

    \bibitem{weatherFNO22pathak}
    Pathak, Jaideep, Shashank Subramanian, Peter Harrington, Sanjeev Raja, Ashesh Chattopadhyay, Morteza Mardani, Thorsten Kurth et al. ``Fourcastnet: A global data-driven high-resolution weather model using adaptive fourier neural operators.'' arXiv preprint arXiv:2202.11214 (2022).

    \bibitem{fno24qin}
    Qin, Shaoxiang, Fuyuan Lyu, Wenhui Peng, Dingyang Geng, Ju Wang, Naiping Gao, Xue Liu, and Liangzhu Leon Wang. ``Toward a Better Understanding of Fourier Neural Operators: Analysis and Improvement from a Spectral Perspective.'' arXiv preprint arXiv:2404.07200 (2024).

    \bibitem{pinn19raissi}
    Raissi, Maziar, Paris Perdikaris, and George E. Karniadakis. ``Physics-informed neural networks: A deep learning framework for solving forward and inverse problems involving nonlinear partial differential equations.'' Journal of Computational physics 378 (2019): 686-707.

    \bibitem{CNO23raonic}
    Raonic, Bogdan, Roberto Molinaro, Tobias Rohner, Siddhartha Mishra, and Emmanuel de Bezenac. ``Convolutional neural operators.'' In ICLR 2023 Workshop on Physics for Machine Learning. 2023

    \bibitem{unet15ron}
    Ronneberger, Olaf, Philipp Fischer, and Thomas Brox. ``U-net: Convolutional networks for biomedical image segmentation.'' In Medical image computing and computer-assisted intervention–MICCAI 2015: 18th international conference, Munich, Germany, October 5-9, 2015, proceedings, part III 18, pp. 234-241. Springer International Publishing, 2015.

    \bibitem{NOMAD22seidman}
    Seidman, Jacob, Georgios Kissas, Paris Perdikaris, and George J. Pappas. ``NOMAD: Nonlinear manifold decoders for operator learning.'' Advances in Neural Information Processing Systems 35 (2022): 5601-5613.

    \bibitem{FFNO21tran}
    Tran, Alasdair, Alexander Mathews, Lexing Xie, and Cheng Soon Ong. ``Factorized fourier neural operators.'' arXiv preprint arXiv:2111.13802 (2021).

    \bibitem{vetterli14signal}
    Vetterli, Martin, Jelena Kovačević, and Vivek K. Goyal. Foundations of signal processing. Cambridge University Press, 2014.

    \bibitem{pdon21wang}
    Wang, Sifan, Hanwen Wang, and Paris Perdikaris. ``Learning the solution operator of parametric partial differential equations with physics-informed DeepONets.'' Science advances 7, no. 40 (2021): eabi8605.

    \bibitem{UFNO22wen}
    Wen, Gege, Zongyi Li, Kamyar Azizzadenesheli, Anima Anandkumar, and Sally M. Benson. ``U-FNO—An enhanced Fourier neural operator-based deep-learning model for multiphase flow.'' Advances in Water Resources 163 (2022): 104180.
\end{thebibliography}
\end{document}